%% file: main.tex
\pdfoutput=1
\RequirePackage[T1]{fontenc}
\documentclass[10pt,journal,compsoc]{IEEEtran}
\usepackage{float}
\input{preamble}
\input{metadata}
\hypersetup{pdftitle={\papertitle}, pdfauthor={Mingqi Yang, Renqi Wu}}

\begin{document}

\title{\papertitle}

\author{\paperauthors%
\IEEEcompsocitemizethanks{%
\IEEEcompsocthanksitem M. Yang and R. Wu are with the South China University
of Technology, Guangzhou, Guangdong, China.\protect\\
E-mail: yangmq@scut.edu.cn, closestick377@gmail.com.
\IEEEcompsocthanksitem This work has been submitted to the IEEE for possible
publication. Copyright may be transferred without notice, after which this
version may no longer be accessible.}}

\markboth{}{}

\IEEEtitleabstractindextext{%
\begin{abstract}
\input{sections/abstract}
\end{abstract}
}

\maketitle
\IEEEdisplaynontitleabstractindextext
\IEEEpeerreviewmaketitle

\input{sections/introduction}
\input{sections/related}
\input{sections/preliminaries}
\input{sections/method}
\input{sections/experiments}
\input{sections/discussion}
\input{sections/conclusion}

\section*{Acknowledgments}
This work was supported by the National Natural Science Foundation of China
(Grant No. 62506134), the Guangdong-Hong Kong-Macao Applied Mathematics
Center Project (Grant No. 2026A1515060008), and the Natural Science
Foundation of Guangdong Province (Grant No. 2026A1515010171).

\bibliographystyle{IEEEtran}
\bibliography{references}

\clearpage
\onecolumn
\begin{center}
{\Large\sffamily Supplementary Material\par}
\end{center}
\medskip
\setcounter{table}{0}
\setcounter{figure}{0}
\setcounter{equation}{0}
\renewcommand{\thetable}{S\arabic{table}}
\renewcommand{\thefigure}{S\arabic{figure}}
\renewcommand{\theequation}{S\arabic{equation}}
\renewcommand{\theHtable}{S\arabic{table}}
\renewcommand{\theHfigure}{S\arabic{figure}}
\renewcommand{\theHequation}{S\arabic{equation}}
\makeatletter
\setlength{\@fptop}{0pt}
\makeatother
\appendices
\input{sections/appendix_method}
\clearpage
\input{sections/appendix_results}

\end{document}

%% file: preamble.tex
\usepackage[nocompress]{cite}
\usepackage{amsmath,amssymb,amsthm}
\usepackage{graphicx}
\usepackage{booktabs}
\usepackage{array}
\usepackage{url}
\usepackage{needspace}
\usepackage[ruled,linesnumbered]{algorithm2e}
\usepackage{xr-hyper}
\usepackage[hidelinks]{hyperref}

\newcommand{\method}{StreamTab}
\newcommand{\parhead}[1]{\par\smallskip\noindent\textbf{#1}\enspace}
\providecommand{\Description}[1]{}

\input{sources/front_stats}

\makeatletter
\let\streamtab@linkfile\hyper@linkfile
\def\hyper@linkfile#1#2#3{%
  \begingroup
  \ifx\\#3\\\else\let\Hy@href@page\@empty\fi
  \streamtab@linkfile{#1}{#2}{#3}%
  \endgroup}
\makeatother

%% file: sources/front_stats.tex
\newcommand{\FrontWithinOneRandom}{seven}

\newcommand{\FrontWinsHistorical}{twelve}

\newcommand{\FrontWinsInductive}{eleven}

\newcommand{\FrontBaselineCount}{21}

%% file: metadata.tex
\newcommand{\papertitle}{StreamTab: Training-Free Temporal Link Prediction with Tabular Foundation Models}
\newcommand{\paperauthors}{Mingqi~Yang and Renqi~Wu}

%% file: sections/abstract.tex
Temporal link prediction uses past interactions to predict whether two
nodes will interact in the future. Temporal graph neural networks achieve
strong results on this task, but they are trained separately for each
graph. How a model should use interaction history also depends on the
candidates it is compared with. A past interaction can distinguish a
recurring link from a random pair that has never interacted, but when the
negative candidates have also interacted before, other signals are
needed. We propose \method{}, a training-free framework built on a frozen
tabular foundation model pretrained only on synthetic tables. \method{}
describes each candidate pair by statistics of its past interactions and
shows the model recent observed interactions and sampled negative pairs
as labeled examples, from which the model scores new candidates. Its
central design, \emph{regime conditioning}, samples the negative examples
according to how the candidates are sampled, so the examples show the
distinction the model must make. As new interactions arrive, \method{}
updates the statistics and refreshes the examples while the model
parameters stay fixed. On thirteen datasets against \FrontBaselineCount{}
temporal graph baselines, \method{} achieves the highest average precision
on \FrontWinsHistorical{} datasets under historical sampling and
\FrontWinsInductive{} under inductive sampling, and ranks first or stays
within one point of the best on \FrontWithinOneRandom{} datasets under
random sampling. With the same features and model, regime conditioning
raises average precision over examples with random negatives by a median
of 12.6 points under historical sampling.

%% file: sections/introduction.tex
\IEEEraisesectionheading{\section{Introduction}\label{sec:intro}}

\IEEEPARstart{T}{emporal} graphs represent entities as nodes and their
interactions as timestamped edges~\cite{holme2012temporal,kazemi2020representation}.
Temporal link prediction uses past interactions to predict whether two
nodes will interact at a given future time~\cite{poursafaei2022towards}.
Applications include anticipating user--item interactions in recommender
systems~\cite{kumar2019predicting,li2020dynamic}, forecasting
communication in social networks~\cite{holme2012temporal,poursafaei2022towards},
and analyzing evolving connections in transportation
networks~\cite{poursafaei2022towards,strohmeier2021crowdsourced}.

Most existing methods learn representations of graph structure and
temporal interaction patterns, and they are trained and selected
separately for each
dataset~\cite{rossi2020temporal,xu2020inductive,wang2021inductive,yu2023towards,tpnet2024,peng2025tidformer,yu2025tami}.
EdgeBank shows that interaction recurrence alone can support competitive
prediction without training~\cite{poursafaei2022towards}, but it turns
history into scores through a fixed rule.

Which historical signals are useful, however, depends on the negative
candidates. Benchmark studies show that replacing random negatives with
historical ones, pairs that have interacted before, changes both
performance and method rankings~\cite{poursafaei2022towards}. A past
interaction can distinguish a recurring link from a random pair that has
never interacted. A historical negative has interacted
before by construction, so recurrence alone cannot tell it from a pair
that interacts again at the query time. These observations raise a
question: \emph{can a predictor learn from labeled examples how to
separate true interactions from the negatives it faces, without parameter
training on graph data?}

Tabular foundation models (TFMs) such as TabPFN offer this possibility
through in-context learning: given a labeled table, they predict labels
for new rows without updating their pretrained
parameters~\cite{muller2022transformers,hollmann2025accurate}.
This capability has been applied to node
classification~\cite{eremeev2025turning}, time-series
forecasting~\cite{hoo2025tables}, static link
prediction~\cite{tfmlinker2026}, and evolving data
streams~\cite{lourenco2026streams}. A TFM reads a table, whereas temporal
link prediction works on a stream of interactions. Using a frozen TFM
therefore means deciding how to describe a candidate pair as a row from
earlier interactions, which labeled rows to show the model, and how to
keep both current as the stream grows.

We propose \method{}, a training-free temporal link prediction framework
built on a frozen TFM. It describes each candidate pair by statistics of
pair recurrence, endpoint activity, and shared neighbors, computed from
earlier interactions and defined even for pairs that have never
interacted. Recent interactions and sampled negative pairs form a labeled
context, from which the TFM scores new candidates. As interactions arrive,
\method{} updates the statistics and periodically refreshes the context.
It uses TabPFN-v2, pretrained only on synthetic tables, and fits no
parameters on graph data.

Our central design is \emph{regime conditioning}: the negatives in the
context are drawn in the same way as the negatives the model will face.
The benchmark protocol, or the system that generates candidates,
specifies this choice. The context then shows the predictor which pairs
it must reject, so one feature map and one frozen predictor serve the random,
historical, and inductive sampling settings of the benchmark by changing
only the context.

We evaluate \method{} on thirteen datasets from the DyGLib
benchmark~\cite{yu2023towards} against \FrontBaselineCount{} temporal
graph baselines. It achieves the highest average precision (AP) on
\FrontWinsHistorical{} datasets under historical sampling and
\FrontWinsInductive{} under inductive sampling. Under
random sampling, it ranks first or stays within one point of the best on
\FrontWithinOneRandom{} datasets. In controlled comparisons with the same
features and frozen predictor, regime conditioning raises AP over a
random-negative context by a median of 12.6 points under historical
sampling. The frozen TFM also outperforms
LightGBM~\cite{ke2017lightgbm} trained on the same features and examples in 35 of 39
comparisons.

Our contributions are summarized as follows:
\begin{itemize}
    \item \textbf{Training-free temporal link prediction.}
    We introduce a framework that turns temporal interaction histories
    into tabular representations and recent labeled contexts, enabling a
    frozen TFM to follow an evolving graph without parameter training on
    graph data.
    \item \textbf{Regime conditioning.}
    We draw the context negatives in the same way as the evaluation
    candidates, so one frozen predictor and feature map serve random,
    historical, and inductive sampling by changing only the context. We show
    that changing the negative distribution changes the posterior and can
    reverse the ranking of candidates.
    \item \textbf{Empirical evidence.}
    On thirteen datasets, \method{} ranks first in most historical- and
    inductive-sampling comparisons against \FrontBaselineCount{} temporal
    graph baselines. Controlled comparisons show substantial gains from regime
    conditioning and an advantage over LightGBM in most comparisons with
    the same inputs. We also examine the accuracy and evaluation cost of
    reusing context across batches.
\end{itemize}
The code is available at \url{https://github.com/qslim/StreamTab}.

%% file: sections/related.tex
\section{Related Work}
\label{sec:related}

\subsection{Temporal Graph Learning}
Temporal graph learning builds representations of evolving structure and
behavior from interaction histories~\cite{kazemi2020representation}.
Memory-based models update node states as interactions
arrive~\cite{kumar2019predicting,rossi2020temporal}. Temporal attention
aggregates timestamped neighborhoods~\cite{xu2020inductive}, propagation
carries information across successive events~\cite{gravina2024long}, and
causal walks and mixer models capture structural and temporal patterns in
these histories~\cite{wang2021inductive,cong2023we}. Link-wise methods
make the relationship between the two candidate nodes part of the
representation, drawing on shared neighborhoods, relative structure, and
the pair's own interaction
history~\cite{yu2023towards,tpnet2024,peng2025tidformer,yu2025tami}.
DyGFormer, for example, encodes neighbor co-occurrence, and TPNet derives
relative encodings from temporal walk matrices. Other designs aggregate a
node's entire temporal neighborhood while avoiding repeated computation
for online inference~\cite{zheng2023tapgnn}, emphasize neighbors with
periodic interaction patterns~\cite{wu2026grpci}, or handle continuous-time
and discrete-time dynamic graphs in one model~\cite{xu2025unidyg}. These
models are trained on each target graph.

Several lines of work reduce this dependence on per-graph training.
EdgeBank scores a pair by a fixed rule over a memory of past edges and is
competitive without any training~\cite{poursafaei2022towards}. Base3
interpolates EdgeBank with node popularity and a lightweight memory
module; it also needs no training and improves on existing baselines
under harder negative sampling~\cite{kondrup2025base3}. TGTalker
prompts large language models with recent temporal neighbors and a few
examples to predict links without training or
fine-tuning~\cite{huang2025tgtalker}. Transfer methods train temporal
graph models on source graphs and apply them to unseen ones. MiNT
pretrains on up to 64 transaction networks and predicts network growth on
unseen networks in a zero-shot setting~\cite{shamsi2024mint}, and
Chatterjee et al.~\cite{chatterjee2025transfer} add a structural mapping
module that initializes the memory of new nodes, so that a trained TGN can
predict links on a disjoint graph without fine-tuning. \method{} also
predicts on a new graph without training on it. Its predictor is
pretrained only on synthetic tables and scores candidate pairs from
tabular history statistics and labeled context examples.

\subsection{Tabular Foundation Models}
Tabular foundation models predict labels for new rows from a labeled
table supplied as
context~\cite{muller2022transformers,hollmann2025accurate,qu2025tabicl}.
Applying them to graphs requires making structure available in tabular
form. G2T-FM constructs tabular node representations for node-level
prediction~\cite{eremeev2025turning}, and GraphPFN adds graph aggregation
layers to a TFM and pretrains it on synthetic
graphs~\cite{eremeev2025graphpfn}. For link
prediction, TFMLinker combines topology-aware link representations with a
prototype-augmented context and fine-tunes the framework, including the
TFM, across static graphs before predicting on new
ones~\cite{tfmlinker2026}.

Temporal extensions of TFMs encode time in the input features or in the
pretraining prior. TabPFN-TS turns forecasting into tabular regression with temporal
features~\cite{hoo2025tables}, and Drift-Resilient TabPFN learns
distribution shifts from a prior with evolving causal
mechanisms~\cite{helli2024drift}. \method{} applies a frozen TFM to temporal
link prediction by describing each candidate pair with the interaction history
available at the query time. It supplies new evidence by updating these
features and the context as interactions arrive.

\subsection{Context Construction and Negative Sampling}
For an in-context predictor, the context decides which labeled evidence is
available. Retrieval and local-context methods select examples near a
query or route queries to relevant parts of the
data~\cite{ma2025tabdpt,thomas2024retrieval,xu2025mixture}, and LoCalPFN
and MixturePFN combine this selection with task-specific fine-tuning. For
evolving data streams, Louren\c{c}o et
al.~\cite{lourenco2026streams} keep long- and short-term memories so that
a frozen TabPFN sees both class coverage and recent evidence.

When temporal link prediction is performed in context, the context also
determines which pairs represent the negative class. Negative mining chooses harder negatives to
train temporal graph models~\cite{chen2025curriculum}, and benchmarks use
historical and inductive negatives to reveal differences that random
negatives hide~\cite{poursafaei2022towards}. \method{} uses this choice at inference
time: regime conditioning draws the context negatives from the sampling
setting under evaluation, so the frozen predictor sees examples of the
pairs it must reject.

%% file: sections/preliminaries.tex
\needspace{4\baselineskip}
\section{Preliminaries}
\label{sec:prelim}

\subsection{Temporal Link Prediction}
A temporal graph records interactions as an ordered event sequence
$\mathcal{E}=(e_1,\ldots,e_N)$, where $e_i=(u_i,v_i,t_i)$ denotes an
interaction from source node $u_i$ to destination node $v_i$ at time $t_i$,
with $t_1\leq\cdots\leq t_N$~\cite{yu2023towards}.
A node pair may interact repeatedly, and new nodes may appear as the
sequence advances. Following the benchmark protocol, the stream is
processed in chronological batches $\mathcal{B}_1,\mathcal{B}_2,\ldots$ of
at most $b$ events. Queries in batch $\mathcal{B}_l$ are answered from the
history $\mathcal{H}_l=\mathcal{B}_1\cup\cdots\cup\mathcal{B}_{l-1}$, and
the batch joins the history once it has been scored.

Given $\mathcal{H}_l$ and a candidate pair $(u,v)$ at query time $t$,
temporal link prediction assigns a score $s(u,v,t)$ indicating how likely
the candidate is to interact at that time. Observed interactions are
positive examples; evaluation compares their scores with those of sampled
negative pairs at the same query
times~\cite{poursafaei2022towards,yu2023towards}.

\subsection{Negative Sampling}
\label{sec:negative-sampling}
Negative sampling determines the alternatives against which observed
interactions are evaluated. In a temporal graph, a pair that has interacted
before can still be a negative candidate when it does not interact during
the current evaluation batch. Recognizing that a pair has interacted is
therefore different from predicting another interaction at the query time.

We consider three settings introduced for
temporal link prediction~\cite{poursafaei2022towards}.
\emph{Random sampling} retains the source node and query time of an
observed interaction and draws a destination uniformly from the candidate
destination set. \emph{Historical sampling} draws from pairs observed
before the time interval covered by the current evaluation batch but absent
within that interval.
\emph{Inductive sampling} further restricts this historical pool to pairs
that had not been observed by a reference cutoff before the evaluation
period. This pair-based setting
differs from new-node evaluation, which concerns nodes unseen during training.
We call $r\in\{\text{random},\text{historical},\text{inductive}\}$ the
\emph{sampling regime}. It specifies which alternatives an observed
interaction is compared with. It is part of the task specification, given
by the benchmark protocol or by the system that generates candidates.

\subsection{Tabular Foundation Models and In-Context Learning}
Tabular foundation models such as TabPFN and TabICL predict labels for
new rows from a labeled table supplied as context, without updating their
pretrained parameters~\cite{hollmann2025accurate,qu2025tabicl}. They
acquire this ability through the pretraining of prior-data fitted networks
(PFNs), which are trained on many sampled tasks to predict held-out labels
from labeled examples~\cite{muller2022transformers}.

For binary classification, let
$C=\{(x_j,y_j)\}_{j=1}^{m}$ be a labeled context, with
$x_j\in\mathbb{R}^{d}$ and $y_j\in\{0,1\}$.
A query row $x\in\mathbb{R}^{d}$ has an unknown label. With pretrained
parameters $\theta$ fixed, the model produces a positive-class score
\begin{equation}
    s=f_\theta(x\mid C).
\end{equation}
The context and query share the same feature columns, and the model
predicts the query label from the labeled examples. Changing the context
changes the task-specific evidence supplied to the predictor.

%% file: sections/method.tex
\section{\method}
\label{sec:method}

\begin{figure*}[!t]
\centering
\includegraphics[width=\linewidth]{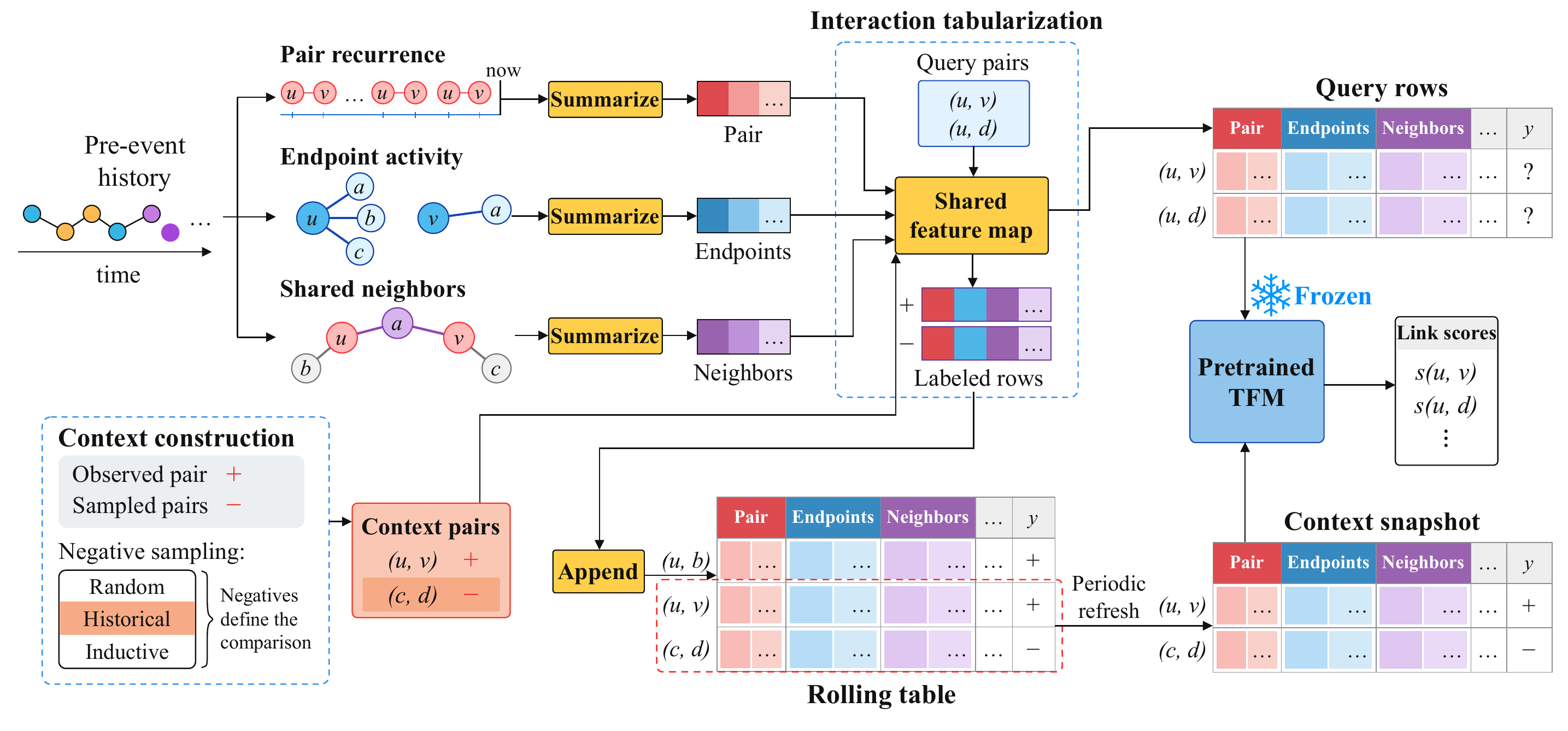}
\caption{Overview of \method{}.
Interaction histories provide tabular features for context examples and
query pairs. Context negatives follow the sampling regime, and
a frozen TFM scores query pairs using recent labeled examples.}
\Description{The upper branches summarize pair recurrence, endpoint
activity, and shared neighbors from pre-event history. A shared feature
map produces query rows and labeled context rows in the same feature
columns. The lower-left selector offers random, historical, and inductive
negative sampling, with historical sampling and its negative pair
highlighted in orange. Labeled rows are appended to a rolling table.
A dashed outline marks recent rows that periodically refresh the context
snapshot. Vertically aligned query and context tables feed a frozen
pretrained tabular foundation model, which outputs candidate link scores.}
\label{fig:overview}
\end{figure*}

\method{} turns temporal link prediction into in-context classification
over interaction histories (Figure~\ref{fig:overview}). Its inputs are the
observed event stream and the sampling regime $r$, and its output is a
score for each candidate pair at its query time. \method{} first
summarizes the past interactions of a pair and the activity of its
endpoints as a tabular feature row (Section~\ref{sec:tabularization}).
Observed interactions and sampled alternatives then supply positive and
negative examples in the same feature space, and the most recent
examples form the context (Section~\ref{sec:context}). A frozen TFM
scores new candidates from this context without updating its parameters,
and both the statistics and the context are updated as new interactions
arrive (Section~\ref{sec:streaming}).

\subsection{Temporal Interaction Tabularization}
\label{sec:tabularization}

A tabular predictor needs one representation for both past examples and
new candidates. For a pair $(u,v)$, the relevant history includes its own
interactions, the activity of each endpoint, and their shared neighbors.
We summarize these aspects as feature columns, and the TFM infers how
they relate to interaction labels from the context. Let $S_l$ summarize
the history $\mathcal{H}_l$ defined in Section~\ref{sec:prelim}. For a
candidate $(u,v)$ queried at time $t$ within batch $\mathcal{B}_l$, its
representation is
\begin{equation}
    x(u,v,t)=\phi(u,v,t\mid S_l)\in\mathcal X,
    \label{eq:tabularization}
\end{equation}
where $\mathcal X$ is a 34-column feature space allowing missing entries.
All candidates of a batch, observed or sampled, are described from the
same state $S_l$, and a row keeps these values when it later serves as
a context example.

Recurrence statistics are easier to interpret together with endpoint
activity. Consider two candidate pairs with the same interaction count and
the same time since their last interaction. For the first pair, both
endpoints have been largely inactive. For the second, both endpoints have
kept interacting with other partners. Pair count and recency treat these
two histories alike. Endpoint activity separates general inactivity from
activity directed elsewhere, and neighborhood overlap separates pairs
that share recent partners from pairs with separate circles of partners.

Table~\ref{tab:feature-groups} organizes this evidence into six blocks.
Pair statistics describe recurrence and age, and source and destination
statistics describe activity, partner diversity, and interaction gaps.
The recent partner-novelty rate measures the fraction of recent
interactions that involved a previously unseen partner. Popularity
columns place endpoint activity relative to the stream, and neighborhood
columns capture mutual appearances in recent partner histories and shared
neighbors.

\begin{table}[!t]
\caption{Historical statistics used to represent a candidate interaction.
Source and destination features are computed separately for each endpoint role.}
\label{tab:feature-groups}
\centering
\begin{tabular}{@{}lrp{0.57\linewidth}@{}}
\toprule
Group & Features & Historical information \\
\midrule
Pair & 7 & Recency, total and windowed counts, time since first interaction, mean recurrence gap,
and elapsed time relative to that gap. \\
Source & 9 & Recency, total and windowed activity, distinct partners,
recent partner novelty, and interaction-gap statistics. \\
Destination & 9 & The corresponding statistics in the destination role. \\
Popularity & 4 & Destination activity shares at two scales, their log
ratio, and the source activity share. \\
Neighborhood & 4 & Endpoint appearances in recent partner histories, common-neighbor
count, and neighborhood Jaccard similarity. \\
Global & 1 & Time since the last event in the history. \\
\bottomrule
\end{tabular}
\end{table}

The same two pairs can also have different recurrence rhythms. Let
$\tau_{uv}$ be the last interaction time of a pair and $\bar\delta_{uv}$
its mean gap over recent inter-interaction intervals. Elapsed
time relative to this rhythm is
\begin{equation}
    \rho_{uv}(t)=\min\!\left\{
        \frac{t-\tau_{uv}}{\bar\delta_{uv}+\epsilon},\rho_{\max}\right\},
    \label{eq:relative-recurrence}
\end{equation}
where $\epsilon>0$ stabilizes the denominator and $\rho_{\max}$ caps
extreme values. The quantity is missing when no positive mean gap is
available. With this ratio, the same elapsed time can mark a long break for one
pair and a routine interval for another.

Activity windows are measured in numbers of events, so their widths have
the same meaning across datasets with different timestamp units. Let
$n_l=|\mathcal H_l|$ be the number of events in the history. The activity
of a pair over the last $w$ events of this history is
\begin{equation}
    c_{uv}^{(w)}(l)=\sum_{j=\max(1,n_l-w+1)}^{n_l}
        \mathbf{1}\{(u_j,v_j)=(u,v)\}.
    \label{eq:window-count}
\end{equation}
Pair and endpoint counts use windows at several scales, which separate
sustained activity from a recent burst. Selected counts and time intervals use a $\log(1+z)$
transformation. Undefined quantities, such as the recurrence gap of an
unseen pair, are missing, and the activity shares of unseen endpoints are
zero. The same feature map therefore applies when new nodes or pairs
enter the stream, without learned identity embeddings.
Appendix~\ref{app:implementation} gives the remaining feature settings.
These columns cover edge recurrence and neighbor co-occurrence, the
signals used by EdgeBank and DyGFormer~\cite{poursafaei2022towards,yu2023towards},
and the context determines how they are combined.

\subsection{Context Construction and Regime Conditioning}
\label{sec:context}

The feature map describes the past of a candidate. The context shows how
that past separates observed interactions from alternatives, so the
choice of alternatives defines the prediction task. In a sparse graph,
uniformly sampled pairs usually have no history, and a previous
interaction is already a strong signal. Historical negatives have a
history by construction, and against them the predictor must recognize
whether a familiar pair is active now. With random negatives in the
context, the model is asked to rank familiar alternatives after being
shown how to reject unfamiliar ones.

\method{} therefore draws its context negatives according to the sampling
regime $r$, a design we call \emph{regime conditioning}. Let $\nu_r$ denote
the context negative generator. Each observed interaction
$e_i=(u_i,v_i,t_i)\in\mathcal{B}_l$ supplies one positive example and $g$
sampled negative examples, forming a labeled block
\begin{equation}
    \begin{aligned}
    D_i^{(r)}=\bigl[&(x(u_i,v_i,t_i),1),\\
    &(x(\tilde u_{i,1},\tilde v_{i,1},t_i),0),\ldots,\\
    &(x(\tilde u_{i,g},\tilde v_{i,g},t_i),0)\bigr],
    \end{aligned}
    \label{eq:context-block}
\end{equation}
where each $(\tilde u_{i,j},\tilde v_{i,j})$ is drawn by $\nu_r$ using only
$\mathcal{H}_l$. The generator mirrors the three regimes:
\begin{itemize}
    \item \emph{Random.} Keep $u_i$ and draw a destination uniformly from
    the distinct destinations observed so far.
    \item \emph{Historical.} Draw uniformly from distinct previously
    observed pairs, excluding $(u_i,v_i)$.
    \item \emph{Inductive.} Draw from observed pairs whose first
    occurrence follows the reference cutoff, excluding $(u_i,v_i)$.
\end{itemize}
The generator approximates the evaluation distribution of regime $r$
from the stream alone and does not read the evaluation candidates.
Appendix~\ref{app:context-sampling} gives the sampling details. Blocks are
appended to a bounded rolling table $\mathcal{T}$, so the table holds
labeled examples in the order their events occurred.

The choice of negatives changes the task itself. Fix the query times and
their histories, and keep the feature map fixed. Let $p_+(x)$ and
$p_-^{(r)}(x)$ be the positive and negative feature densities under regime
$r$, and let $\pi\in(0,1)$ be the positive proportion. We keep $p_+$ and
$\pi$ the same across regimes. The resulting classification task has the
posterior
\begin{equation}
    \eta_r(x)=P_r(Y=1\mid X=x)
    =\frac{\pi p_+(x)}{\pi p_+(x)+(1-\pi)p_-^{(r)}(x)},
    \label{eq:regime-posterior}
\end{equation}
where $Y=1$ for observed interactions and $Y=0$ for sampled pairs. This is
the density-ratio view used in noise-contrastive
estimation~\cite{gutmann2010noise}. Up to the class prior, the odds of a
positive label equal $p_+(x)/p_-^{(r)}(x)$. The negative distribution therefore
determines how strongly a given history supports a positive label.

The change can also reorder candidates. For two regimes $r$ and $r'$,
\begin{equation}
    \operatorname{logit}\eta_r(x)-\operatorname{logit}\eta_{r'}(x)
    =\log\frac{p_-^{(r')}(x)}{p_-^{(r)}(x)}.
    \label{eq:regime-ordering}
\end{equation}
The correction depends on the candidate, so a change of regime can
reverse the ordering of two candidates, which a shift of the decision
threshold cannot do (Appendix~\ref{app:regime-ordering}). Applying the
posterior of another regime also adds an expected Kullback--Leibler
divergence to the log loss, by the standard decomposition of the
logarithmic score~\cite{gneiting2007strictly}
(Appendix~\ref{app:regime-analysis}). This analysis concerns the ideal
posteriors. In practice, the frozen TFM estimates the feature--label
relationship from a finite table of recent positives and selected
alternatives.

\subsection{Streaming In-Context Prediction}
\label{sec:streaming}

New interactions matter in two ways: they change the features of later
candidates, and they supply labeled examples for later queries. \method{}
updates the history state $S$ and the rolling table $\mathcal{T}$ after
every batch and refreshes the context every $R$ batches, which we call
the refresh interval. Conditioning the TFM on a new context is the costly
step, so refreshing it less often lets all queries of a cycle be scored
together against the same context.

Events preceding the evaluation period initialize $S$ and $\mathcal{T}$
with the same batch procedure. A refresh cycle $q$ spans at most $R$
consecutive batches. At its start, the context is the most recent part of
the table,
\begin{equation}
    C_q^{(r)}=\operatorname{Tail}_k\!\left(\mathcal{T}\right),
    \label{eq:context-snapshot}
\end{equation}
where $\operatorname{Tail}_k$ selects the last at most $k$ rows. Positive
and negative examples share this budget, and $g$ sets their ratio. For a
candidate $(u,v)$ queried at time $t$ in a batch of cycle $q$, the score is
\begin{equation}
    s(u,v,t)=f_\theta\!\left(
        \phi(u,v,t\mid S_l)\mid C_{q}^{(r)}\right).
    \label{eq:streaming-score}
\end{equation}
Algorithm~\ref{alg:streamtab} summarizes the procedure. A batch's labeled
blocks enter $\mathcal{T}$ only after the batch has been scored, and its
events then enter $S$. Later batches see them through updated features
and, from the next cycle on, through the context. A larger context size $k$ covers a longer span of history, and a larger
$R$ spreads the conditioning cost over more queries while delaying the use
of new examples. Query features still include every preceding batch
throughout the cycle. Section~\ref{sec:experimental-setup} gives the
values of $k$, $R$, and $g$.

\begin{algorithm}[!t]
\caption{\method{} streaming prediction}
\label{alg:streamtab}
\small
\KwIn{Event stream in batches $\mathcal{B}_1,\mathcal{B}_2,\ldots$; query
candidates; frozen TFM $f_\theta$; negative generator $\nu_r$; context
size $k$; refresh interval $R$; negatives per event $g$.}
Initialize $S$ and $\mathcal{T}$ from the batches preceding evaluation\;
\ForEach{refresh cycle $q$ of at most $R$ batches}{
  $C_q^{(r)}\leftarrow\operatorname{Tail}_k(\mathcal{T})$; condition $f_\theta$ on $C_q^{(r)}$\;
  \ForEach{batch $\mathcal{B}_l$ in the cycle}{
    Score query candidates with $f_\theta(\phi(\cdot\mid S)\mid C_q^{(r)})$\;
    \ForEach{event $e_i\in\mathcal{B}_l$}{
      Build $D_i^{(r)}$ from $S$ and $\nu_r$ (Equation~\ref{eq:context-block})\;
    }
    Add the events of $\mathcal{B}_l$ to $S$ and append the blocks to $\mathcal{T}$\;
  }
}
\end{algorithm}

The history indices occupy $O(N)$ space after $N$ events, alongside the
bounded rolling table. Each evaluation event requires feature rows for its
query candidates and its $g$ context negatives. Over $n_b$ evaluation
batches, the TFM is conditioned $\lceil n_b/R\rceil$ times, and each
conditioning scores all queries of its cycle against the same $k$
context rows.

%% file: sections/experiments.tex
\section{Experiments}
\label{sec:experiments}

We evaluate \method{} on thirteen temporal graph datasets. We first compare
it with published temporal graph methods, then examine where its accuracy
comes from, and finally study how context size and refresh trade accuracy
for cost over the stream. The experiments address the following
\textbf{Research Questions (RQs)}:
\begin{itemize}
    \item \textbf{RQ1} (Section~\ref{sec:rq1}): How does \method{} perform
    against existing temporal graph methods?
    \item \textbf{RQ2} (Section~\ref{sec:rq2}): How does regime conditioning
    affect link prediction?
    \item \textbf{RQ3} (Section~\ref{sec:rq3}): How does the frozen TFM compare
    with other predictors given the same inputs?
    \item \textbf{RQ4} (Section~\ref{sec:rq4}): How do context size and refresh
    frequency affect prediction accuracy and evaluation cost?
    \item \textbf{RQ5} (Section~\ref{sec:rq5}): How does context refresh affect
    prediction over time?
\end{itemize}

\subsection{Experimental Setup}
\label{sec:experimental-setup}

\parhead{Datasets.}
We use the thirteen datasets of the DyGLib
benchmark~\cite{poursafaei2022towards,yu2023towards}: Wikipedia, Reddit,
MOOC, LastFM, Enron, Social Evo., UCI, Flights, Can. Parl., US Legis., UN Trade,
UN Vote, and Contact. They span social platforms, learning and
music-listening logs, communication and proximity networks, air traffic,
and political and trade networks.
Table~\ref{tab:dataset-statistics} lists their statistics. Test surprise is
the fraction of test interactions whose node pair was not observed in the
training or validation period. It is close to zero on Social Evo., UN Vote,
and UN Trade, where nearly every test pair has occurred before, and
highest on MOOC and UCI, at 0.796 and 0.739.

\begin{table}[!t]
\caption{Statistics of the temporal graph datasets. Interactions include
repeated occurrences of a pair; unique pairs count distinct
source--destination pairs. Bip. indicates a bipartite graph;
Surprise is the fraction of test interactions involving pairs absent from
the training and validation periods.}
\label{tab:dataset-statistics}
\centering
\scriptsize
\setlength{\tabcolsep}{1.8pt}
\input{tables/dataset_statistics}
\end{table}

\parhead{Baselines.}
We compare with 21 temporal graph baselines: memory-based models
(JODIE~\cite{kumar2019predicting}, DyRep~\cite{trivedi2019dyrep},
TGN~\cite{rossi2020temporal}, CTAN~\cite{gravina2024long}); methods built on
temporal walks and structural encodings (CAWN~\cite{wang2021inductive},
NAT~\cite{luo2022neighborhood}, PINT~\cite{souza2022provably},
TPNet~\cite{tpnet2024}, CNE-N~\cite{cheng2024coneighbor},
TAWRMAC~\cite{farokhi2025tawrmac}); attention, mixer, and sequence models
(TGAT~\cite{xu2020inductive}, TCL~\cite{wang2021tcl},
GraphMixer~\cite{cong2023we}, DyGFormer~\cite{yu2023towards},
FreeDyG~\cite{tian2024freedyg}, DyGMamba~\cite{ding2025dygmamba},
RepeatMixer~\cite{zou2024repeat}, TIDFormer~\cite{peng2025tidformer},
TAMI~\cite{yu2025tami}); and the training-free
EdgeBank~\cite{poursafaei2022towards}. TAMI contributes two variants,
TAMI-GraphMixer and TAMI-DyGFormer.

\parhead{Evaluation protocol.}
Following DyGLib~\cite{yu2023towards}, we split each dataset
chronologically at the 70th and 85th percentiles of its timestamps and
evaluate the transductive setting under the three negative sampling
regimes of Poursafaei et al.~\cite{poursafaei2022towards}, with one
negative per positive interaction, fixed across runs. We report average
precision (AP) and the area under the ROC curve (AUC), computed within
chronological batches of 200 positive interactions and averaged over
batches. \method{} scores each batch using the events of all earlier batches, including
the training and validation periods, and takes the sampling regime of the
evaluation as input when it draws context negatives
(Section~\ref{sec:negative-sampling}); the baselines are trained with
random negatives. The new-node evaluation keeps the test queries that
involve nodes held out from training; \method{} answers them from the same
event prefix, which includes earlier interactions of those nodes.

\parhead{Implementation details.}
\method{} uses the released pretrained TabPFN-v2
classifier~\cite{hollmann2025accurate} with four ensemble members and keeps
its parameters fixed. Candidates are described by the 34 historical
features of Section~\ref{sec:tabularization}, computed from event
identities and timestamps alone. One configuration serves every dataset
and regime: a context of $k=8{,}192$ rows with $g=1$ negative per positive,
refreshed every $R=4$ batches. We use this configuration for all datasets and
regimes, and Section~\ref{sec:rq4} shows that nearby settings give similar
accuracy. We report the mean and standard deviation
over three runs (seeds 0--2). Experiments run on a server with Intel Xeon
Gold 6430 processors and NVIDIA RTX 4090 GPUs.

\input{sources/rq1_stats}
\subsection{Comparison with Temporal Graph Methods (RQ1)}
\label{sec:rq1}

Each sampling regime asks a different question. Random negatives ask
whether a predictor can tell a true interaction from an arbitrary pair.
Historical and inductive negatives ask whether it can tell a currently
active pair from pairs that have interacted before. We report all three
regimes and a new-node evaluation.
Appendix~\ref{app:full-results} of the supplementary material reports standard deviations and AUC.

\begin{table*}[!t]
\caption{Full-test AP (\%) under historical negative sampling. Bold and
underline mark the best and second-best means at the displayed precision;
ties share the same mark. Wins / 13 counts first places.
A dash indicates an unavailable result.}
\label{tab:rq1-historical}
\centering
\scriptsize
\setlength{\tabcolsep}{2.4pt}
\input{tables/rq1_historical_wide}
\end{table*}

\begin{table*}[!t]
\caption{Full-test AP (\%) under inductive negative sampling. Conventions
follow Table~\ref{tab:rq1-historical}.}
\label{tab:rq1-inductive}
\centering
\scriptsize
\setlength{\tabcolsep}{2.4pt}
\input{tables/rq1_inductive_wide}
\end{table*}

\parhead{Historical negative sampling.}
Historical negatives have interacted before, so a past interaction alone
cannot identify the true one. Table~\ref{tab:rq1-historical} reports mean AP under historical sampling.
\method{} ranks first on \RqOneHistWins{} of
the thirteen datasets, with a median
margin of \RqOneHistMedian{} percentage points over the strongest published
result. Its largest gain is \RqOneHistMaxGain{} points over
\RqOneHistMaxGainBase{} on \RqOneHistMaxGainData{}, and DyGMamba leads on
Can. Parl. by \RqOneHistCanParlMargin{} points. The context of \method{}
contains previously observed pairs as negatives, so the predictor learns
from the same kind of comparison that the test uses. Historical sampling
is the setting in which \method{} is strongest.

\parhead{Inductive negative sampling.}
Inductive negatives are pairs that first appeared after the reference
cutoff, so the predictor must also judge recently formed pairs. As Table~\ref{tab:rq1-inductive} shows,
\method{} ranks first in mean AP on \RqOneIndWins{} datasets,
with a median margin of \RqOneIndMedian{} points and a largest gain of
\RqOneIndMaxGain{} points over \RqOneIndMaxGainBase{} on
\RqOneIndMaxGainData{}. DyGMamba and TPNet lead on Can. Parl. and
US Legis. As new pairs enter the stream, \method{} adds them to its context
and describes them with the same historical features. These results come
from a model whose parameters never change on the target graph; only its
features and context follow the stream.

\parhead{Random negative sampling.}
Under random sampling, most negatives have no shared history, so
recurrence alone already separates many of them from true interactions.
Table~\ref{tab:rq1-random} in the supplementary material reports mean AP under random sampling.
\method{} ranks first on \RqOneRandWins{} datasets and is within one point
of the best result on \RqOneRandWithinOne{}.
Its largest gaps occur on MOOC, UCI, and Can. Parl., the three datasets
with the highest test surprise. On these datasets, many true interactions are
also new pairs, so pair recurrence separates them poorly from random
negatives. Under random sampling, the gaps of \method{} are concentrated
on datasets where true interactions often lack a pair history.

\parhead{New-node evaluation.}
Table~\ref{tab:rq1-newnode} in the supplementary material reports the
new-node results for each dataset. These queries involve nodes held out
from training. \method{} computes its features from the available
interaction history, so it handles such nodes without learned node
embeddings. Under historical sampling, it ranks first on
\RqOneNewHistWins{} datasets with a median margin of \RqOneNewHistMedian{}
points, close to the \RqOneHistMedian{} points on the full test set, and
gains 22.29 points on Flights and 18.35 points on UN Vote. Under inductive
sampling, it ranks first on \RqOneNewIndWins{} datasets, with gains of
19.80 points on UN Vote and 17.05 points on Flights. Under random sampling,
it ranks first on \RqOneNewRandWins{} datasets and stays within one point
of the best result on \RqOneNewRandWithinOne{}. The strength of \method{}
under historical sampling thus carries over to nodes held out from
training. The next section examines how much of these gains comes from the
matched context.

\input{sources/rq2_stats}
\subsection{Effect of Regime Conditioning (RQ2)}
\label{sec:rq2}

Regime conditioning is the only part of \method{} that changes across
sampling regimes, so we measure its effect by changing the context
negatives alone. We compare regime-conditioned and random-negative contexts
on all thirteen datasets, and cross the three context strategies with the
three test settings on UCI, MOOC, LastFM, and Enron. Features, context
size, and refresh interval stay fixed.

\begin{figure*}[!t]
\centering
\includegraphics[width=\linewidth]{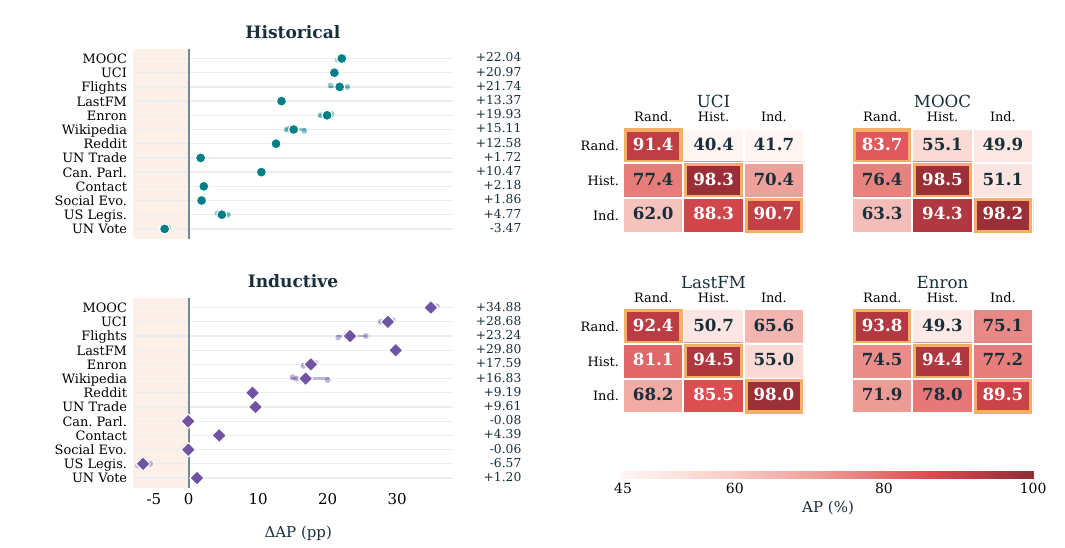}
\caption{Effect of regime conditioning. Left: AP differences between regime-matched and random-negative
contexts, in percentage points (pp); large markers show means over three
runs, small markers show individual runs, and bars show their range.
Right: mean AP (\%) for each combination of test sampling (rows) and
context sampling (columns). Outlined cells mark the best context for each test setting.}
\label{fig:rq2}
\end{figure*}

\parhead{Gains from context conditioning.}
Under historical and inductive sampling, the negatives have interacted
before, and a context of random negatives contains few such pairs. Drawing
context negatives from the evaluation regime raises AP by a median of
\RqTwoHistMedian{} points under historical sampling and \RqTwoIndMedian{}
points under inductive sampling (Figure~\ref{fig:rq2}, left). The largest
gains are \RqTwoHistMax{} and \RqTwoIndMax{} points, both on
\RqTwoHistMaxData{}. Conditioning improves \RqTwoHistPositive{} and
\RqTwoIndPositive{} of the thirteen datasets in the two settings. With random-negative
context, \method{} beats the strongest published result on
\RqTwoHistRandomBeats{} datasets in each setting. With matched context, it
does so on \RqTwoHistMatchedBeats{} and \RqTwoIndMatchedBeats{} datasets.
The matched context is the main source of the lead of \method{} in these
two settings.

\parhead{Context rankings across tasks.}
We next cross the three context strategies with the three test settings.
In all twelve rows, the best context is the one that matches the test
setting (Figure~\ref{fig:rq2}, right). On UCI, random context exceeds historical context by 51 points under
random testing, and historical context exceeds random context by 21 points
under historical testing. MOOC, LastFM, and Enron show the same pattern.
Each context teaches the predictor a different comparison. One frozen model
can therefore serve all three regimes by changing only its context
negatives.

\parhead{Asymmetric transfer.}
Inductive negatives are a subset of the historical pool
(Section~\ref{sec:negative-sampling}), so the two contexts may not
transfer equally well. Under inductive testing, historical context trails the
matched context by \RqTwoHistCtxGapMin{}--\RqTwoHistCtxGapMax{} points,
and random context trails it by \RqTwoRandCtxGapMin{}--\RqTwoRandCtxGapMax{}
points. Under
historical testing, inductive context does worse than random context on
UCI, MOOC, and LastFM. On MOOC, it reaches 51.15 AP, while historical
context reaches 98.48. Historical context covers the pairs that inductive
testing uses. Inductive context lacks the older pairs that historical
testing also draws on. Historical context thus transfers much better, but
it does not replace the matched context.

\subsection{Effect of the Predictor (RQ3)}
\label{sec:rq3}

The context decides what the predictor sees. We next ask how much the
frozen TFM adds when it reads this context. TabPFN,
LightGBM~\cite{ke2017lightgbm}, histogram gradient boosting (HGB) from
scikit-learn~\cite{pedregosa2011scikit}, and TabICL~\cite{qu2025tabicl} receive the same
historical features, labeled context, and refresh schedule, with
$k=8{,}192$ and $R=4$. The two tree models are refitted whenever the
context changes, and the two TFMs keep their pretrained parameters. We
compare TabPFN and LightGBM on all thirteen datasets and all four
predictors on six datasets. Table~\ref{tab:rq3} reports four of these
datasets, Table~\ref{tab:rq3-all} gives the full results, and
Appendix~\ref{app:predictor-settings} lists the predictor settings.

\begin{table}[!t]
\caption{Predictor comparison with the same inputs. Top: TabPFN versus
LightGBM across thirteen datasets. Wins count datasets where TabPFN
achieves higher mean AP than LightGBM; $\Delta$AP denotes TabPFN minus
LightGBM, measured in percentage points (pp). Below: AP (\%) on four datasets
(mean $\pm$ standard deviation over three runs). Bold marks the best mean.}
\label{tab:rq3}
\centering
\input{tables/rq3_predictors_revised}
\end{table}

Tree models refitted at every refresh are the natural alternative to a
frozen TFM, since they also learn from the context alone. TabPFN has
higher AP than LightGBM in 35 of the 39 dataset--regime comparisons. Its median gains are 0.64, 0.13, and 2.28 points under
random, historical, and inductive sampling. The largest gains are 8.95
points on UCI under random sampling and 10.48 points on UN Trade under
inductive sampling. LightGBM leads in the other four comparisons, by up to
3.89 points on UN Vote under historical sampling. Under historical
sampling, TabPFN, LightGBM, and HGB stay within 0.57 points of each other
on each of the six datasets. In this setting, the features and the
matched context already carry most of the signal. The frozen TFM adds the
most under inductive sampling and the least under historical sampling,
and it outperforms trees refitted at every refresh in most settings
without being fitted to the target graph.

To check whether this advantage is specific to TabPFN, we also evaluate
TabICL. It exceeds both tree models in 15 of the 18 comparisons on the six
datasets. It is the best predictor in four comparisons, where it leads
TabPFN by at most 0.2 points. TabPFN leads in the other fourteen by a
median of 0.30 points, and its largest leads are 7.95 points on UCI under
random sampling and 4.01 points on Enron under inductive sampling. Both
pretrained TFMs show this advantage.

\input{sources/rq4_stats}
\subsection{Performance and Evaluation Cost (RQ4)}
\label{sec:rq4}

\begin{figure}[!t]
\centering
\includegraphics[width=\columnwidth]{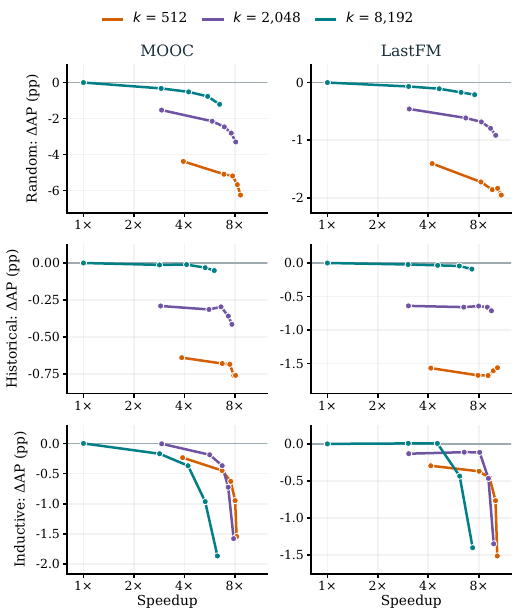}
\caption{AP change and evaluation speedup across fifteen context
configurations on MOOC and LastFM. Each line represents a context size $k$, with points for
$R=1,4,8,16,32$. AP differences (percentage points) and evaluation
speedups are relative to $k=8{,}192$, $R=1$ for the same dataset and
sampling regime, using means over three runs.}
\label{fig:rq4}
\end{figure}

\method{} has two core parameters. The context size $k$ is the number of
labeled rows the TFM reads for each prediction. The refresh
interval $R$ is the number of batches between two context rebuilds. More
rows and more frequent rebuilds give the predictor more and fresher
examples, but they make evaluation slower. We evaluate all fifteen combinations of $k\in\{512,2048,8192\}$ and $R\in\{1,4,8,16,32\}$ under the
three sampling settings on six datasets. Figure~\ref{fig:rq4} shows MOOC
and LastFM, and Figure~\ref{fig:rq4-all} shows all six datasets. Speedups
use mean full-test evaluation times over three runs, relative to
$k=8{,}192$ and $R=1$.

Query features keep updating with every batch while the labeled rows stay
fixed, so the same context can be reused across several batches. On MOOC
and LastFM, refreshing every four batches instead of every batch makes
evaluation 2.8--3.1$\times$ faster and removes about two thirds of the
evaluation time. AP drops by at most 0.33 points. With $R=32$, evaluation
becomes 6.0--7.6$\times$ faster, and AP falls by up to 1.21 points under
random and historical sampling and by up to 1.86 points under inductive
sampling. A refresh interval of four keeps nearly all of the accuracy of
per-batch refresh at about a third of its cost.

Reducing the context size also saves time, but it removes examples. At
$R=4$, using 512 rows instead of 8,192 makes evaluation 2.4$\times$ faster
on MOOC and 2.6--2.7$\times$ faster on LastFM. The largest AP loss is 4.77
points on MOOC under random sampling. We also compare the two choices at
similar runtimes. On MOOC under random sampling, $k=8{,}192$ with $R=32$
reaches 82.80 AP in 75.75 seconds, and $k=2{,}048$ with $R=4$ reaches 81.85
AP in 83.93 seconds. The larger context with less frequent refresh is
more accurate under random and historical sampling on both datasets. Under
inductive sampling, the smaller context refreshed more often leads by 1.68
points on MOOC and 1.29 points on LastFM. The inductive test compares
recently formed pairs, so fresh examples matter more there than a large
context.

The default $k=8{,}192$, $R=4$ serves every dataset and regime. On MOOC and
LastFM, it stays within 0.33 points of the best of the fifteen
configurations in every setting. The other four datasets in
Figure~\ref{fig:rq4-all} show gaps below 0.4 points, except on UCI and on
Enron under inductive sampling, where smaller contexts do better by up to
2.92 points. A single shared configuration is therefore close to the best
per-task choice, and it does not need to be tuned for each dataset or
regime.

\input{sources/rq5_stats}
\subsection{Context Refresh over Time (RQ5)}
\label{sec:rq5}

Refresh also changes which examples the predictor sees as the test period
goes on. To measure this effect, we compare $R=4$ with a fixed initial
context on UCI, MOOC, LastFM, and Enron, using $k=8{,}192$ under all three
sampling settings. In both variants, historical statistics and query
features keep updating, and only the labeled examples differ.

\begin{table}[!t]
\caption{AP (\%) with periodic refresh and a fixed initial context
(mean $\pm$ standard deviation over three matched runs).
Refresh uses $R=4$; both variants use $k=8{,}192$.
$\Delta$AP is the mean paired difference (refresh minus fixed), in
percentage points.}
\label{tab:rq5}
\centering
\footnotesize
\setlength{\tabcolsep}{4pt}
\input{tables/rq5_refresh}
\end{table}

A fixed context ages as the stream moves on, so we first ask whether
refreshing it changes accuracy. Refresh raises AP in ten of the twelve
comparisons in Table~\ref{tab:rq5}, and LastFM under historical sampling
is unchanged. The largest gains are 2.31 and 2.20
points on UCI and MOOC under random sampling and 2.05 points on LastFM
under inductive sampling, and each of them is positive in all three runs.
Under historical sampling, the gains on UCI and MOOC are small, 0.16 and
0.03 points. Refresh also makes results more stable. The standard
deviation across runs is lower with refresh in nine of the twelve
comparisons, and on UCI under random sampling it falls from 1.68 to 0.27.

Across all thirteen datasets, the median gains are \RqFiveAllRandMedian{},
\RqFiveAllHistMedian{}, and \RqFiveAllIndMedian{} points under random,
historical, and inductive sampling. The largest gain is \RqFiveAllIndMax{}
points on \RqFiveAllIndMaxData{} under inductive sampling, and refresh
lowers historical-sampling AP on \RqFiveAllNegAData{} and
\RqFiveAllNegBData{} by \RqFiveAllNegA{} and \RqFiveAllNegB{} points.
Refresh therefore helps in most settings. Its effect under historical
sampling is small and can be negative on the UN datasets.

\begin{figure}[!t]
\centering
\includegraphics[width=\columnwidth]{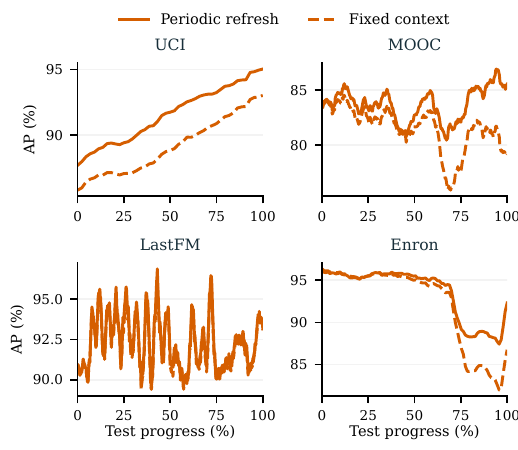}
\caption{AP over the test period under random sampling. Curves show mean AP over three runs with matched seeds, smoothed with
a centered 25-batch moving average and shorter windows at the boundaries.}
\label{fig:rq5}
\end{figure}

Figure~\ref{fig:rq5} shows how the two variants differ over the test
period, and Appendix~\ref{app:refresh} gives the historical and inductive
curves. On UCI, both variants improve over time, and the refreshed context
leads from early on. On MOOC, the gap grows in the later part of the
stream. On Enron, both curves drop near the end, and the fixed context
drops more. On LastFM, the two curves stay close. On MOOC and Enron, the
fixed context falls further behind as the test period moves away from the
initial context, so the value of refresh grows over time on these
datasets.

%% file: tables/dataset_statistics.tex
\begin{tabular}{@{}llrrrcr@{}}
\toprule
Dataset & Domain & Nodes & Interactions & Unique pairs & Bip. & Surprise \\
\midrule
Wikipedia & Social & 9,227 & 157,474 & 18,257 & Yes & 0.291 \\
Reddit & Social & 10,984 & 672,447 & 78,516 & Yes & 0.111 \\
MOOC & Interaction & 7,144 & 411,749 & 178,443 & Yes & 0.796 \\
LastFM & Interaction & 1,980 & 1,293,103 & 154,993 & Yes & 0.419 \\
Enron & Social & 184 & 125,235 & 3,125 & No & 0.269 \\
Social Evo. & Proximity & 74 & 2,099,519 & 4,486 & No & 0.002 \\
UCI & Social & 1,899 & 59,835 & 20,296 & No & 0.739 \\
Flights & Transport & 13,169 & 1,927,145 & 395,072 & No & 0.177 \\
Can. Parl. & Politics & 734 & 74,478 & 51,331 & No & 0.567 \\
US Legis. & Politics & 225 & 60,396 & 26,423 & No & 0.340 \\
UN Trade & Economics & 255 & 507,497 & 36,182 & No & 0.026 \\
UN Vote & Politics & 201 & 1,035,742 & 31,516 & No & 0.004 \\
Contact & Proximity & 692 & 2,426,279 & 79,530 & No & 0.102 \\
\bottomrule
\end{tabular}

%% file: sources/rq1_stats.tex
\newcommand{\RqOneHistWins}{twelve}
\newcommand{\RqOneHistWithinOne}{twelve}
\newcommand{\RqOneHistMedian}{4.01}
\newcommand{\RqOneHistMaxGain}{19.21}
\newcommand{\RqOneHistMaxGainData}{Flights}
\newcommand{\RqOneHistMaxGainBase}{CTAN}
\newcommand{\RqOneHistAucWins}{eleven}
\newcommand{\RqOneHistAucWithinOne}{twelve}
\newcommand{\RqOneHistAucMedian}{2.84}
\newcommand{\RqOneHistAucMaxGain}{15.98}
\newcommand{\RqOneHistAucMaxGainData}{Flights}
\newcommand{\RqOneHistAucMaxGainBase}{CTAN}
\newcommand{\RqOneHistwikipediaMargin}{0.74}
\newcommand{\RqOneHistwikipediaBase}{CTAN}
\newcommand{\RqOneHistredditMargin}{4.01}
\newcommand{\RqOneHistredditBase}{CTAN}
\newcommand{\RqOneHistmoocMargin}{3.11}
\newcommand{\RqOneHistmoocBase}{TIDFormer}
\newcommand{\RqOneHistlastfmMargin}{5.80}
\newcommand{\RqOneHistlastfmBase}{TIDFormer}
\newcommand{\RqOneHistenronMargin}{6.21}
\newcommand{\RqOneHistenronBase}{RepeatMixer}
\newcommand{\RqOneHistSocialEvoMargin}{1.71}
\newcommand{\RqOneHistSocialEvoBase}{FreeDyG}
\newcommand{\RqOneHistuciMargin}{4.38}
\newcommand{\RqOneHistuciBase}{PINT}
\newcommand{\RqOneHistFlightsMargin}{19.21}
\newcommand{\RqOneHistFlightsBase}{CTAN}
\newcommand{\RqOneHistCanParlMargin}{1.75}
\newcommand{\RqOneHistCanParlBase}{DyGMamba}
\newcommand{\RqOneHistUSLegisMargin}{0.71}
\newcommand{\RqOneHistUSLegisBase}{TPNet}
\newcommand{\RqOneHistUNtradeMargin}{6.12}
\newcommand{\RqOneHistUNtradeBase}{TPNet}
\newcommand{\RqOneHistUNvoteMargin}{5.57}
\newcommand{\RqOneHistUNvoteBase}{EdgeBank}
\newcommand{\RqOneHistContactsMargin}{1.44}
\newcommand{\RqOneHistContactsBase}{TPNet}
\newcommand{\RqOneIndWins}{eleven}
\newcommand{\RqOneIndWithinOne}{eleven}
\newcommand{\RqOneIndMedian}{3.24}
\newcommand{\RqOneIndMaxGain}{12.85}
\newcommand{\RqOneIndMaxGainData}{UN Vote}
\newcommand{\RqOneIndMaxGainBase}{TAMI-GraphMixer}
\newcommand{\RqOneIndAucWins}{ten}
\newcommand{\RqOneIndAucWithinOne}{eleven}
\newcommand{\RqOneIndAucMedian}{7.55}
\newcommand{\RqOneIndAucMaxGain}{18.70}
\newcommand{\RqOneIndAucMaxGainData}{LastFM}
\newcommand{\RqOneIndAucMaxGainBase}{CTAN}
\newcommand{\RqOneIndwikipediaMargin}{3.24}
\newcommand{\RqOneIndwikipediaBase}{CTAN}
\newcommand{\RqOneIndredditMargin}{2.94}
\newcommand{\RqOneIndredditBase}{TAMI-GraphMixer}
\newcommand{\RqOneIndmoocMargin}{5.82}
\newcommand{\RqOneIndmoocBase}{TAMI-DyGFormer}
\newcommand{\RqOneIndlastfmMargin}{6.94}
\newcommand{\RqOneIndlastfmBase}{TAMI-GraphMixer}
\newcommand{\RqOneIndenronMargin}{0.54}
\newcommand{\RqOneIndenronBase}{TAMI-GraphMixer}
\newcommand{\RqOneIndSocialEvoMargin}{0.05}
\newcommand{\RqOneIndSocialEvoBase}{TAMI-DyGFormer}
\newcommand{\RqOneInduciMargin}{4.92}
\newcommand{\RqOneInduciBase}{TIDFormer}
\newcommand{\RqOneIndFlightsMargin}{5.23}
\newcommand{\RqOneIndFlightsBase}{TAMI-GraphMixer}
\newcommand{\RqOneIndCanParlMargin}{15.16}
\newcommand{\RqOneIndCanParlBase}{DyGMamba}
\newcommand{\RqOneIndUSLegisMargin}{7.96}
\newcommand{\RqOneIndUSLegisBase}{TPNet}
\newcommand{\RqOneIndUNtradeMargin}{7.60}
\newcommand{\RqOneIndUNtradeBase}{TAMI-GraphMixer}
\newcommand{\RqOneIndUNvoteMargin}{12.85}
\newcommand{\RqOneIndUNvoteBase}{TAMI-GraphMixer}
\newcommand{\RqOneIndContactsMargin}{1.32}
\newcommand{\RqOneIndContactsBase}{TAMI-DyGFormer}
\newcommand{\RqOneRandWins}{two}
\newcommand{\RqOneRandWithinOne}{seven}
\newcommand{\RqOneRandMedian}{$-$0.80}
\newcommand{\RqOneRandMaxGain}{2.62}
\newcommand{\RqOneRandMaxGainData}{UN Vote}
\newcommand{\RqOneRandMaxGainBase}{TPNet}
\newcommand{\RqOneRandAucWins}{two}
\newcommand{\RqOneRandAucWithinOne}{nine}
\newcommand{\RqOneRandAucMedian}{$-$0.78}
\newcommand{\RqOneRandAucMaxGain}{2.63}
\newcommand{\RqOneRandAucMaxGainData}{UN Vote}
\newcommand{\RqOneRandAucMaxGainBase}{TPNet}
\newcommand{\RqOneRandwikipediaMargin}{0.69}
\newcommand{\RqOneRandwikipediaBase}{TPNet}
\newcommand{\RqOneRandredditMargin}{0.32}
\newcommand{\RqOneRandredditBase}{FreeDyG}
\newcommand{\RqOneRandmoocMargin}{12.71}
\newcommand{\RqOneRandmoocBase}{TPNet}
\newcommand{\RqOneRandlastfmMargin}{2.23}
\newcommand{\RqOneRandlastfmBase}{TIDFormer}
\newcommand{\RqOneRandenronMargin}{0.66}
\newcommand{\RqOneRandenronBase}{TIDFormer}
\newcommand{\RqOneRandSocialEvoMargin}{0.80}
\newcommand{\RqOneRandSocialEvoBase}{FreeDyG}
\newcommand{\RqOneRanduciMargin}{6.00}
\newcommand{\RqOneRanduciBase}{TIDFormer}
\newcommand{\RqOneRandFlightsMargin}{0.40}
\newcommand{\RqOneRandFlightsBase}{DyGMamba}
\newcommand{\RqOneRandCanParlMargin}{15.24}
\newcommand{\RqOneRandCanParlBase}{DyGMamba}
\newcommand{\RqOneRandUSLegisMargin}{1.64}
\newcommand{\RqOneRandUSLegisBase}{TPNet}
\newcommand{\RqOneRandUNtradeMargin}{1.28}
\newcommand{\RqOneRandUNtradeBase}{TPNet}
\newcommand{\RqOneRandUNvoteMargin}{2.62}
\newcommand{\RqOneRandUNvoteBase}{TPNet}
\newcommand{\RqOneRandContactsMargin}{0.01}
\newcommand{\RqOneRandContactsBase}{TAWRMAC}
\newcommand{\RqOneNewHistWins}{nine}
\newcommand{\RqOneNewHistMedian}{3.42}
\newcommand{\RqOneNewHistWithinOne}{nine}
\newcommand{\RqOneNewHistMaxGap}{19.29}
\newcommand{\RqOneNewHistMaxGapData}{Can. Parl.}
\newcommand{\RqOneNewIndWins}{four}
\newcommand{\RqOneNewIndMedian}{$-$3.34}
\newcommand{\RqOneNewIndWithinOne}{four}
\newcommand{\RqOneNewIndMaxGap}{27.96}
\newcommand{\RqOneNewIndMaxGapData}{LastFM}
\newcommand{\RqOneNewRandWins}{three}
\newcommand{\RqOneNewRandMedian}{$-$0.89}
\newcommand{\RqOneNewRandWithinOne}{seven}
\newcommand{\RqOneNewRandMaxGap}{28.73}
\newcommand{\RqOneNewRandMaxGapData}{Can. Parl.}

%% file: tables/rq1_historical_wide.tex
\begin{tabular}{@{}lcccccccccccccc@{}}
\toprule
Method & Wiki. & Reddit & MOOC & LastFM & Enron & S.Evo & UCI & Flights & CanP. & USLeg. & UNTr. & UNVo. & Cont. & Wins / 13 \\
\midrule
JODIE & 83.01 & 80.03 & 78.94 & 74.35 & 69.85 & 87.44 & 75.24 & 66.48 & 51.79 & 51.71 & 61.39 & 70.02 & 95.31 & 0 \\
DyRep & 79.93 & 79.83 & 75.60 & 74.92 & 71.19 & 93.29 & 55.10 & 67.61 & 63.31 & 86.88 & 59.19 & 69.30 & 96.39 & 0 \\
TGN & 86.86 & 81.22 & 87.06 & 76.87 & 73.91 & 94.45 & 80.43 & 66.70 & 68.42 & 74.00 & 58.44 & 69.37 & 93.05 & 0 \\
CTAN & \underline{95.91} & \underline{89.77} & 67.73 & 82.29 & 77.24 & --- & 76.62 & \underline{75.75} & 89.28 & 84.20 & 67.38 & 74.75 & 97.86 & 0 \\
CAWN & 71.21 & 80.82 & 74.05 & 69.86 & 64.73 & 85.53 & 65.30 & 64.72 & 66.53 & 68.82 & 55.71 & 51.26 & 84.16 & 0 \\
NAT & 68.36 & 80.58 & 85.42 & 78.75 & 72.90 & 91.29 & 75.38 & 64.74 & 77.72 & 91.12 & 78.65 & 71.39 & 96.84 & 0 \\
PINT & 91.33 & 83.93 & 84.80 & 84.95 & 85.41 & 96.73 & \underline{93.96} & 66.82 & 63.84 & 73.58 & 70.07 & 71.79 & 97.61 & 0 \\
TPNet & 81.55 & 81.02 & 92.69 & 87.74 & 80.79 & 96.80 & 86.34 & 69.10 & 86.61 & \underline{94.55} & \underline{85.22} & 74.68 & \underline{98.02} & 0 \\
TAWRMAC & 90.16 & 84.36 & 88.93 & 82.69 & 80.23 & --- & 88.29 & 68.57 & 76.60 & 91.17 & 67.99 & 74.67 & 97.81 & 0 \\
TGAT & 87.38 & 79.55 & 82.19 & 71.59 & 64.07 & 95.01 & 68.27 & 72.38 & 67.13 & 62.14 & 55.74 & 52.96 & 96.05 & 0 \\
TCL & 89.05 & 77.14 & 77.06 & 59.30 & 70.66 & 94.74 & 80.25 & 70.68 & 65.93 & 80.53 & 55.90 & 52.30 & 93.86 & 0 \\
GraphMixer & 90.90 & 78.44 & 77.77 & 72.47 & 77.98 & 94.93 & 84.11 & 71.47 & 74.34 & 81.65 & 57.05 & 51.20 & 93.36 & 0 \\
DyGFormer & 82.23 & 81.57 & 85.85 & 81.57 & 75.63 & 97.38 & 82.17 & 66.59 & 97.00 & 85.30 & 64.41 & 60.84 & 97.57 & 0 \\
FreeDyG & 91.59 & 85.67 & 86.71 & 79.71 & 78.87 & \underline{97.79} & 86.10 & --- & --- & --- & --- & --- & --- & 0 \\
DyGMamba & 81.77 & 81.80 & 85.89 & 83.02 & 77.77 & 97.35 & 81.03 & 67.80 & \textbf{99.77} & 82.15 & 65.10 & 61.07 & 97.61 & 1 \\
RepeatMixer & 91.02 & 84.44 & 94.24 & 88.41 & \underline{88.19} & --- & 87.23 & --- & --- & --- & --- & --- & --- & 0 \\
TIDFormer & 91.21 & 83.27 & \underline{95.37} & \underline{88.68} & 81.59 & 97.39 & 89.57 & --- & --- & --- & --- & --- & --- & 0 \\
TAMI-GraphMixer & 90.97 & 82.56 & 83.61 & 80.23 & 81.68 & 96.80 & 87.69 & 73.01 & 78.81 & 84.56 & 69.74 & 70.90 & 93.30 & 0 \\
TAMI-DyGFormer & 82.38 & 85.18 & 86.26 & 83.40 & 81.02 & 97.56 & 85.89 & 67.77 & \underline{98.96} & 86.10 & 65.16 & 81.72 & 97.72 & 0 \\
EdgeBank & 73.35 & 73.59 & 60.71 & 73.03 & 76.53 & 80.57 & 65.50 & 70.53 & 63.84 & 63.22 & 81.32 & \underline{84.89} & 88.81 & 0 \\
\midrule
\method{} & \textbf{96.65} & \textbf{93.78} & \textbf{98.48} & \textbf{94.48} & \textbf{94.40} & \textbf{99.50} & \textbf{98.34} & \textbf{94.96} & 98.02 & \textbf{95.26} & \textbf{91.34} & \textbf{90.46} & \textbf{99.46} & 12 \\
\bottomrule
\end{tabular}

%% file: tables/rq1_inductive_wide.tex
\begin{tabular}{@{}lcccccccccccccc@{}}
\toprule
Method & Wiki. & Reddit & MOOC & LastFM & Enron & S.Evo & UCI & Flights & CanP. & USLeg. & UNTr. & UNVo. & Cont. & Wins / 13 \\
\midrule
JODIE & 75.65 & 86.98 & 65.23 & 62.67 & 68.96 & 89.82 & 65.99 & 69.07 & 48.42 & 50.27 & 60.42 & 67.79 & 93.43 & 0 \\
DyRep & 70.21 & 86.30 & 61.66 & 64.41 & 67.79 & 93.28 & 54.79 & 70.57 & 58.61 & 83.44 & 60.19 & 67.53 & 94.18 & 0 \\
TGN & 85.62 & 88.10 & 77.50 & 65.95 & 70.89 & 95.13 & 70.94 & 71.09 & 65.34 & 67.57 & 61.04 & 67.63 & 90.18 & 0 \\
CTAN & \underline{94.15} & 90.99 & 64.93 & 80.06 & 72.02 & --- & 66.25 & 76.20 & 87.42 & 79.13 & 68.98 & 72.85 & 97.31 & 0 \\
CAWN & 74.06 & 91.67 & 73.51 & 67.48 & 75.15 & 88.32 & 64.61 & 69.18 & 67.75 & 65.81 & 62.54 & 52.19 & 89.31 & 0 \\
NAT & 53.85 & 81.71 & 77.28 & 64.02 & 70.34 & 92.12 & 57.55 & 59.16 & 78.03 & \underline{88.40} & 78.38 & 72.53 & 91.13 & 0 \\
PINT & 82.39 & 87.59 & 75.70 & 72.24 & 80.23 & 97.15 & 83.61 & 61.99 & 59.15 & 67.78 & 69.72 & 68.37 & 94.05 & 0 \\
TPNet & 79.35 & 88.19 & 88.18 & 77.99 & 75.36 & 97.35 & 77.26 & 64.78 & 85.59 & \textbf{91.05} & 86.61 & 75.05 & 95.84 & 1 \\
TAWRMAC & 89.55 & 90.80 & 83.42 & 72.57 & 78.58 & --- & 81.46 & 65.62 & 69.87 & 84.02 & 60.64 & 69.63 & 96.01 & 0 \\
TGAT & 87.00 & 89.59 & 75.95 & 71.13 & 63.94 & 94.84 & 68.67 & 75.48 & 68.82 & 61.91 & 60.61 & 52.89 & 94.35 & 0 \\
TCL & 86.76 & 87.45 & 74.65 & 58.21 & 71.29 & 94.90 & 76.01 & 74.62 & 65.85 & 78.15 & 61.06 & 50.62 & 91.35 & 0 \\
GraphMixer & 88.59 & 85.26 & 74.27 & 68.12 & 75.01 & 94.72 & 80.10 & 74.87 & 69.48 & 79.63 & 60.15 & 51.60 & 90.87 & 0 \\
DyGFormer & 78.29 & 91.11 & 81.24 & 73.97 & 77.41 & 97.68 & 72.25 & 70.92 & 95.44 & 81.25 & 55.79 & 51.91 & 94.75 & 0 \\
FreeDyG & 90.05 & 90.74 & 83.01 & 72.19 & 77.81 & 97.57 & 82.35 & --- & --- & --- & --- & --- & --- & 0 \\
DyGMamba & 79.86 & 91.15 & 81.11 & 73.63 & 80.86 & 97.68 & 71.95 & 73.79 & \textbf{98.32} & 81.67 & 58.89 & 52.24 & 95.43 & 1 \\
RepeatMixer & 88.98 & 91.76 & 83.68 & 76.34 & 84.17 & --- & 84.96 & --- & --- & --- & --- & --- & --- & 0 \\
TIDFormer & 91.15 & 91.57 & 84.53 & 75.69 & 79.67 & 97.70 & \underline{85.81} & --- & --- & --- & --- & --- & --- & 0 \\
TAMI-GraphMixer & 93.89 & \underline{96.19} & 79.82 & \underline{91.06} & \underline{88.95} & 96.09 & 84.12 & \underline{93.64} & 70.94 & 83.31 & \underline{87.73} & \underline{79.53} & 96.12 & 0 \\
TAMI-DyGFormer & 78.96 & 94.37 & \underline{92.39} & 74.03 & 86.23 & \underline{97.76} & 80.13 & 75.55 & \underline{97.25} & 81.31 & 68.01 & 78.19 & \underline{98.47} & 0 \\
EdgeBank & 80.63 & 85.48 & 49.43 & 75.49 & 73.89 & 83.69 & 57.43 & 81.08 & 62.16 & 64.74 & 72.97 & 66.30 & 85.20 & 0 \\
\midrule
\method{} & \textbf{97.39} & \textbf{99.13} & \textbf{98.21} & \textbf{98.00} & \textbf{89.49} & \textbf{97.81} & \textbf{90.73} & \textbf{98.87} & 83.16 & 83.09 & \textbf{95.33} & \textbf{92.38} & \textbf{99.79} & 11 \\
\bottomrule
\end{tabular}

%% file: sources/rq2_stats.tex
\newcommand{\RqTwoMismatchMedian}{27.3}
\newcommand{\RqTwoMismatchMin}{2.4}
\newcommand{\RqTwoMismatchMax}{51.1}
\newcommand{\RqTwoMatchedBest}{twelve}
\newcommand{\RqTwoHistMedian}{12.6}
\newcommand{\RqTwoHistPositive}{twelve}
\newcommand{\RqTwoHistMax}{22.0}
\newcommand{\RqTwoHistMaxData}{MOOC}
\newcommand{\RqTwoHistRandomBeats}{two}
\newcommand{\RqTwoHistMatchedBeats}{twelve}
\newcommand{\RqTwoHistWorst}{3.5}
\newcommand{\RqTwoHistWorstData}{UN Vote}
\newcommand{\RqTwoIndMedian}{9.6}
\newcommand{\RqTwoIndPositive}{ten}
\newcommand{\RqTwoIndMax}{34.9}
\newcommand{\RqTwoIndMaxData}{MOOC}
\newcommand{\RqTwoIndRandomBeats}{two}
\newcommand{\RqTwoIndMatchedBeats}{eleven}
\newcommand{\RqTwoIndWorst}{6.6}
\newcommand{\RqTwoIndWorstData}{US Legis.}
\newcommand{\RqTwoHistCtxGapMin}{2.4}
\newcommand{\RqTwoHistCtxGapMax}{12.5}
\newcommand{\RqTwoRandCtxGapMin}{17.6}
\newcommand{\RqTwoRandCtxGapMax}{34.9}

%% file: tables/rq3_predictors_revised.tex
\begingroup\footnotesize
\setlength{\tabcolsep}{3pt}
\renewcommand{\arraystretch}{1.12}
\begin{tabular}{@{}lcc@{}}
\toprule
Sampling regime & Wins / 13 & Median $\Delta$AP (pp) \\
\midrule
Random & 12 & $+0.64$ \\
Historical & 10 & $+0.13$ \\
Inductive & 13 & $+2.28$ \\
\bottomrule
\end{tabular}
\par\smallskip
\begin{tabular}{@{}lcccc@{}}
\toprule
Random & UCI & MOOC & LastFM & Enron \\
\midrule
TabPFN & $\mathbf{91.44}_{\pm 0.27}$ & $\mathbf{83.68}_{\pm 0.08}$ & $\mathbf{92.37}_{\pm 0.03}$ & $\mathbf{93.83}_{\pm 0.03}$ \\
LightGBM & $82.49_{\pm 0.12}$ & $82.44_{\pm 0.11}$ & $91.75_{\pm 0.03}$ & $93.19_{\pm 0.15}$ \\
HGB & $83.81_{\pm 0.69}$ & $81.75_{\pm 0.05}$ & $91.61_{\pm 0.02}$ & $92.84_{\pm 0.09}$ \\
TabICL & $83.49_{\pm 0.50}$ & $82.43_{\pm 0.16}$ & $92.36_{\pm 0.02}$ & $93.51_{\pm 0.07}$ \\
\bottomrule
\end{tabular}
\par\smallskip
\begin{tabular}{@{}lcccc@{}}
\toprule
Historical & UCI & MOOC & LastFM & Enron \\
\midrule
TabPFN & $98.34_{\pm 0.01}$ & $\mathbf{98.48}_{\pm 0.01}$ & $\mathbf{94.48}_{\pm 0.02}$ & $\mathbf{94.40}_{\pm 0.09}$ \\
LightGBM & $98.13_{\pm 0.02}$ & $98.40_{\pm 0.01}$ & $94.17_{\pm 0.01}$ & $94.27_{\pm 0.08}$ \\
HGB & $98.08_{\pm 0.04}$ & $98.37_{\pm 0.01}$ & $94.05_{\pm 0.02}$ & $94.17_{\pm 0.09}$ \\
TabICL & $\mathbf{98.36}_{\pm 0.01}$ & $98.46_{\pm 0.01}$ & $94.21_{\pm 0.03}$ & $93.62_{\pm 0.08}$ \\
\bottomrule
\end{tabular}
\par\smallskip
\begin{tabular}{@{}lcccc@{}}
\toprule
Inductive & UCI & MOOC & LastFM & Enron \\
\midrule
TabPFN & $90.73_{\pm 0.09}$ & $98.21_{\pm 0.05}$ & $\mathbf{98.00}_{\pm 0.01}$ & $\mathbf{89.49}_{\pm 0.89}$ \\
LightGBM & $90.19_{\pm 0.24}$ & $98.05_{\pm 0.03}$ & $97.60_{\pm 0.02}$ & $82.25_{\pm 0.98}$ \\
HGB & $89.77_{\pm 0.35}$ & $97.97_{\pm 0.03}$ & $97.55_{\pm 0.01}$ & $82.15_{\pm 0.40}$ \\
TabICL & $\mathbf{90.78}_{\pm 0.19}$ & $\mathbf{98.23}_{\pm 0.01}$ & $97.86_{\pm 0.02}$ & $85.48_{\pm 0.38}$ \\
\bottomrule
\end{tabular}
\par\smallskip
\endgroup

%% file: sources/rq4_stats.tex
\newcommand{\RqFourCells}{eighteen}
\newcommand{\RqFourDefaultSpeedMin}{2.3}
\newcommand{\RqFourDefaultSpeedMax}{3.1}
\newcommand{\RqFourDefaultLossMedian}{0.02}
\newcommand{\RqFourDefaultLossMax}{1.51}
\newcommand{\RqFourDefaultLossMaxData}{UCI}
\newcommand{\RqFourDefaultLossMaxRegime}{inductive}
\newcommand{\RqFourDefaultWithin}{seventeen}
\newcommand{\RqFourLongSpeedMin}{3.6}
\newcommand{\RqFourLongSpeedMax}{7.6}
\newcommand{\RqFourLongLossMax}{1.77}
\newcommand{\RqFourSmallSpeedMin}{1.7}
\newcommand{\RqFourSmallSpeedMax}{2.7}
\newcommand{\RqFourSmallLossMedian}{0.57}
\newcommand{\RqFourSmallLossMax}{4.77}
\newcommand{\RqFourSmallLossMaxData}{MOOC}
\newcommand{\RqFourSmallLossMaxRegime}{random}
\newcommand{\RqFourSmallBetterCells}{four}
\newcommand{\RqFourLargeSparseWins}{eleven}
\newcommand{\RqFourLargeSparseRuntimeMin}{0.89}
\newcommand{\RqFourLargeSparseRuntimeMax}{0.99}
\newcommand{\RqFourDefaultBestCells}{one}
\newcommand{\RqFourBestGapMedian}{0.10}
\newcommand{\RqFourBestGapA}{2.92}
\newcommand{\RqFourBestGapAData}{Enron}
\newcommand{\RqFourBestGapARegime}{inductive}
\newcommand{\RqFourBestGapB}{2.72}
\newcommand{\RqFourBestGapBData}{UCI}
\newcommand{\RqFourBestGapBRegime}{inductive}
\newcommand{\RqFourWallMin}{35}
\newcommand{\RqFourWallMinData}{UCI}
\newcommand{\RqFourWallMax}{486}
\newcommand{\RqFourWallMaxData}{LastFM}

%% file: sources/rq5_stats.tex
\newcommand{\RqFiveFourPositive}{eleven}
\newcommand{\RqFiveFourTotal}{twelve}
\newcommand{\RqFiveTopA}{2.31}
\newcommand{\RqFiveTopAData}{UCI}
\newcommand{\RqFiveTopARegime}{random}
\newcommand{\RqFiveTopASeeds}{three}
\newcommand{\RqFiveTopB}{2.20}
\newcommand{\RqFiveTopBData}{MOOC}
\newcommand{\RqFiveTopBRegime}{random}
\newcommand{\RqFiveTopBSeeds}{three}
\newcommand{\RqFiveTopC}{2.05}
\newcommand{\RqFiveTopCData}{LastFM}
\newcommand{\RqFiveTopCRegime}{inductive}
\newcommand{\RqFiveTopCSeeds}{three}
\newcommand{\RqFiveLowA}{0.09}
\newcommand{\RqFiveLowAData}{Enron}
\newcommand{\RqFiveLowARegime}{inductive}
\newcommand{\RqFiveLowASign}{-}
\newcommand{\RqFiveLowB}{0.00}
\newcommand{\RqFiveLowBData}{LastFM}
\newcommand{\RqFiveLowBRegime}{historical}
\newcommand{\RqFiveLowBSign}{+}
\newcommand{\RqFiveEnronHist}{0.76}
\newcommand{\RqFiveAllRandPositive}{ten}
\newcommand{\RqFiveAllRandMedian}{0.46}
\newcommand{\RqFiveAllRandMax}{3.16}
\newcommand{\RqFiveAllRandMaxData}{UN Vote}
\newcommand{\RqFiveAllHistPositive}{eleven}
\newcommand{\RqFiveAllHistMedian}{0.13}
\newcommand{\RqFiveAllHistMax}{1.52}
\newcommand{\RqFiveAllHistMaxData}{Can. Parl.}
\newcommand{\RqFiveAllIndPositive}{eight}
\newcommand{\RqFiveAllIndMedian}{0.71}
\newcommand{\RqFiveAllIndMax}{9.47}
\newcommand{\RqFiveAllIndMaxData}{Can. Parl.}
\newcommand{\RqFiveAllTotal}{39}
\newcommand{\RqFiveAllNegA}{3.78}
\newcommand{\RqFiveAllNegAData}{UN Vote}
\newcommand{\RqFiveAllNegARegime}{historical}
\newcommand{\RqFiveAllNegB}{3.45}
\newcommand{\RqFiveAllNegBData}{UN Trade}
\newcommand{\RqFiveAllNegBRegime}{historical}

%% file: tables/rq5_refresh.tex
\begin{tabular}{@{}llccr@{}}
\toprule
Dataset & Regime & Fixed & Refresh & $\Delta$AP \\
\midrule
UCI & Random & 89.12 $\pm$ 1.68 & 91.44 $\pm$ 0.27 & +2.31 \\
 & Historical & 98.18 $\pm$ 0.01 & 98.34 $\pm$ 0.01 & +0.16 \\
 & Inductive & 90.02 $\pm$ 0.30 & 90.73 $\pm$ 0.09 & +0.71 \\
\addlinespace[2pt]
MOOC & Random & 81.48 $\pm$ 0.15 & 83.68 $\pm$ 0.08 & +2.20 \\
 & Historical & 98.45 $\pm$ 0.00 & 98.48 $\pm$ 0.01 & +0.03 \\
 & Inductive & 97.43 $\pm$ 0.07 & 98.21 $\pm$ 0.05 & +0.78 \\
\addlinespace[2pt]
LastFM & Random & 92.13 $\pm$ 0.10 & 92.37 $\pm$ 0.03 & +0.24 \\
 & Historical & 94.48 $\pm$ 0.08 & 94.48 $\pm$ 0.02 & +0.00 \\
 & Inductive & 95.95 $\pm$ 0.06 & 98.00 $\pm$ 0.01 & +2.05 \\
\addlinespace[2pt]
Enron & Random & 92.57 $\pm$ 0.15 & 93.83 $\pm$ 0.03 & +1.26 \\
 & Historical & 93.64 $\pm$ 0.31 & 94.40 $\pm$ 0.09 & +0.76 \\
 & Inductive & 89.58 $\pm$ 0.74 & 89.49 $\pm$ 0.89 & $-$0.09 \\
\bottomrule
\end{tabular}

%% file: sections/discussion.tex
\section{Discussion}
\label{sec:discussion}

\parhead{Why does \method{} work?}
\method{} uses a TFM pretrained only on synthetic tables and never updates
it on graph data, so the task has to be specified through its inputs. The
features describe the interaction history of each candidate pair, and the
labeled context shows the predictor which pairs count as negatives. With
random-negative context, \method{} leads on two datasets in each of the
historical and inductive settings. With matched context, it leads on most
datasets in both. The pretrained predictor adds a smaller but consistent
gain: with the same features and context, TabPFN outperforms LightGBM
refitted at every refresh in 35 of 39 comparisons, and TabICL shows the
same advantage. Most of the lead under historical and inductive sampling
thus comes from the matched context. The same frozen model moves to
another regime by rebuilding its context, with no retraining and no change
to the features.

\begin{figure}[!t]
\centering
\includegraphics[width=\columnwidth]{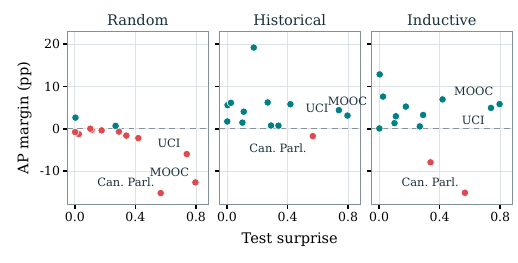}
\caption{AP difference between \method{} and the best published baseline
versus test surprise under three negative sampling regimes. Positive
values favor \method{}; each point represents one dataset.}
\label{fig:surprise}
\end{figure}

\parhead{When does \method{} work best?}
Test surprise measures how many test interactions involve new pairs, so it
shows how much interaction history is available. Figure~\ref{fig:surprise}
plots the margin of \method{} against test surprise under the three
sampling regimes. Under random sampling, the margin falls as surprise
rises, and the large gaps appear on Can. Parl., UCI, and MOOC, where many
test pairs are new. Under historical and inductive sampling, this trend
largely disappears. MOOC and UCI move from the largest gaps to clear
leads, and \method{} leads on most datasets regardless of surprise, with
Can. Parl. as the main exception. When the negatives have also interacted
before, recent changes in a pair's history become the main evidence, even
on graphs where most test pairs are new. On queries with nodes held out
from training, the lead under historical sampling remains, while under
inductive sampling it narrows to four datasets. The history features help
most when the negatives have a history of their own.

\parhead{What does training-free prediction cost?}
\method{} spends no time on training, so its cost lies in inference. With
the shared default, one full pass over the test period takes 35 to 486
seconds on a single GPU across the six datasets of Section~\ref{sec:rq4}.
Context size and refresh interval control this cost, and the default sits
close to the best of the fifteen configurations on almost every task. The
cost can therefore be lowered without tuning for each dataset.

\parhead{Limitations and future work.}
The features use only event identities and timestamps. Node and edge
attributes, and learned structural features, could add evidence for pairs
with little shared history, where \method{} gains least under random
sampling. The context is drawn from one candidate regime at a time.
Systems that mix several candidate sources could use contexts that
combine regimes in the proportions the generator produces. Context
construction also offers room for improvement. Selecting examples that
are relevant to each query and refreshing only the examples that have
become stale could make better use of a fixed context budget. Finally,
larger benchmarks with longer streams~\cite{huang2023temporal} would test how far a frozen
predictor with a bounded context can follow a changing graph.

%% file: sections/conclusion.tex
\section{Conclusion}
\label{sec:conclusion}

\method{} shows that a frozen tabular foundation model, pretrained only on
synthetic tables, can predict future interactions from statistics of past
interactions and a table of labeled examples, without parameter training
on graph data. Sampling the negative examples according to how the
candidates are sampled lets one model serve all three sampling settings,
and \method{} is strongest when the candidates have interacted before.
The frozen model also outperforms tree models refitted on the same
examples in most comparisons, and the examples can be reused across
several batches with little loss in accuracy. These results point to the
construction of labeled examples as a practical way to adapt a pretrained
predictor to evolving graphs. Pairs with little shared history remain the
main challenge, and node and edge attributes, together with examples
selected for each query, are natural ways to address them.

%% file: sections/appendix_method.tex
\section{Implementation Details}
\label{app:implementation}

\subsection{Historical Features}
Activity windows count events, with widths of 1,000 and 10,000 events for
pairs and 100, 1,000, and 10,000 events for endpoints. Each endpoint keeps
its last 16 partners in the corresponding source or destination role for
partner-appearance counts and the partner-novelty rate, and its endpoint
gap statistics use up to 16 inter-interaction intervals in that role. For
neighborhood overlap, we take the partners in the last 32 interactions in
each role and unite the two role-specific sets of each endpoint. The
common-neighbor count and the Jaccard similarity are computed from these
sets.

Pair mean gaps use up to eight recent intervals. In
Equation~\ref{eq:relative-recurrence}, $\epsilon=10^{-9}$ and
$\rho_{\max}=10^6$. Pair recency, total count, 10,000-event count, age,
and mean gap use $\log(1+z)$, and the 1,000-event count and relative
recurrence keep their original scales. In each endpoint block, recency,
total count, 1,000- and 10,000-event counts, and distinct-partner count
use $\log(1+z)$. Endpoint gap means and standard deviations are computed
after applying $\log(1+z)$ to each interval, with the population standard
deviation. The 100-event count, the novelty rate, and the
partner-appearance counts keep their original scales.

Activity shares divide the windowed count by the smaller of the window
width and the number of preceding events, with a minimum denominator of
one. The destination trend is
$\log[(p_{1000}+\epsilon)/(p_{10000}+\epsilon)]$, where $p_w$ is the
destination activity share over window $w$. The global feature is the
log-transformed time since the preceding event. Pair and endpoint blocks
are missing for pairs and nodes without history in the corresponding
role, the activity shares of such endpoints are zero, and the destination
trend is missing. A gap mean needs at least one interval, and an
endpoint gap standard deviation needs at least two. Relative recurrence is
missing when the pair mean gap is not positive. When either endpoint has
no history in its role, the common-neighbor count and the Jaccard
similarity are missing, and the partner-appearance count is zero for a
role with history and missing for a role without history.

\subsection{Context Negative Sampling}
\label{app:context-sampling}
The context generators of Section~\ref{sec:context} use the history state
at the start of the current batch, the same state from which the batch is
scored. For inductive sampling, the reference cutoff is the end of training while
the context is initialized and the end of validation during the test
stream. Context rows keep their values and labels when the cutoff
changes.

Inductive and historical sampling draw uniformly from the eligible
distinct pairs and reject the current positive pair. When no eligible pair
is drawn, inductive sampling falls back to historical sampling and
historical sampling falls back to random sampling. Random sampling keeps the current
source and draws uniformly from the observed destinations. These
rules only define the examples placed in the context; the benchmark
sampler generates the evaluation candidates independently.

\subsection{Predictor Configuration}
\label{app:predictor-settings}
\method{} uses the TabPFN-v2 classifier of the \texttt{tabpfn} package,
version 2.0.9, with four ensemble members. TabICL uses version 2.1.1 of
the \texttt{tabicl} package. In the predictor comparison of
Section~\ref{sec:rq3}, all predictors receive the same feature columns,
context rows, and refresh schedule. LightGBM uses 400 estimators, a
learning rate of 0.05, 63 leaves, and a feature-sampling fraction of 0.9, with eight CPU
threads. HGB uses at most
300 boosting iterations. Other tree-model parameters keep their library
defaults. TabICL receives missing entries replaced by $-1$ in context and
query rows, and the other predictors receive the missing entries directly.
The run seed sets the random state of every predictor. If a context
contains only one class, every query receives a score of 0.5 until a
context with both classes is available.

\section{Analysis of Regime Conditioning}
\label{app:regime-analysis}

\subsection{Log-Loss Cost of a Mismatched Regime}
For a scoring rule $h:\mathcal X\to[0,1]$, the binary log loss on the
sampled classification task $P_r$ of regime $r$ is
\begin{equation}
    \mathcal L_r(h)=\mathbb E_{P_r}
    \bigl[-Y\log h(X)-(1-Y)\log(1-h(X))\bigr].
\end{equation}
Let $\eta_r$ and $\eta_s$ be the posteriors of
Equation~\ref{eq:regime-posterior} under a target regime $r$ and a
reference regime $s$, with feature marginals satisfying
$P_r^X\ll P_s^X$, so that $\eta_s$ is defined almost everywhere under the
target distribution. Then
\begin{equation}
    \mathcal L_r(\eta_s)-\mathcal L_r(\eta_r)
    =\mathbb E_{X\sim P_r^X}\!\left[
    D_{\mathrm{KL}}\!\left(
    \operatorname{Ber}(\eta_r(X))\,\Vert\,
    \operatorname{Ber}(\eta_s(X))\right)\right].
    \label{eq:regime-mismatch}
\end{equation}
The excess loss is zero exactly when the two posteriors agree
$P_r^X$-almost everywhere, and positive otherwise.

\begin{proof}
Condition on $X=x$ and write $a=\eta_r(x)$ and $b=\eta_s(x)$. The
conditional loss of $b$ minus that of $a$ is
\begin{equation*}
    a\log\frac{a}{b}+(1-a)\log\frac{1-a}{1-b}
    =D_{\mathrm{KL}}(\operatorname{Ber}(a)\Vert\operatorname{Ber}(b)).
\end{equation*}
Integrating over $P_r^X$ gives Equation~\ref{eq:regime-mismatch}. The
target Bayes risk is at most $\log 2$, so the identity remains well defined
when the cross-regime risk is infinite. Nonnegativity and the equality
condition follow from those of the KL divergence, with the usual
conventions at probabilities zero and one.
\end{proof}

The identity places the cost of a mismatch in the regions where the two
regimes give the same features different meanings, weighted by how often
these regions occur in the target task.

\subsection{A Candidate-Ordering Example}
\label{app:regime-ordering}
Consider a two-point feature space $X\in\{a,b\}$ with $\pi=1/2$ and the
distributions in Table~\ref{tab:regime-ordering-example}. The positive
distribution is the same in both regimes. Regime $r$ places more negative
mass on $a$, which makes $b$ the stronger positive candidate, and regime
$s$ reverses this comparison. Equation~\ref{eq:regime-posterior} gives
$\eta_r(a)=5/13<5/7=\eta_r(b)$ and $\eta_s(a)=5/7>5/13=\eta_s(b)$.

\begin{table}[!htbp]
\caption{A constructed example in which two negative distributions
reverse the ordering of two candidates.}
\label{tab:regime-ordering-example}
\centering
\begin{tabular}{@{}cccccc@{}}
\toprule
$x$ & $p_+(x)$ & $p_-^{(r)}(x)$ & $p_-^{(s)}(x)$
& $\eta_r(x)$ & $\eta_s(x)$ \\
\midrule
$a$ & $1/2$ & $4/5$ & $1/5$ & $5/13$ & $5/7$ \\
$b$ & $1/2$ & $1/5$ & $4/5$ & $5/7$ & $5/13$ \\
\bottomrule
\end{tabular}
\end{table}

Whether the ordering changes depends on the difference between the
candidate-specific corrections in Equation~\ref{eq:regime-ordering}.
Replacing $p_-^{(s)}=(1/5,4/5)$ by $(7/10,3/10)$, for example, changes the
posteriors under $s$ to $(5/12,5/8)$ but keeps the ordering of $r$. A
mismatch in probabilities and a change in ordering are therefore distinct
effects. The log-loss identity measures the first, and AP and AUC evaluate
the second.

%% file: sections/appendix_results.tex
\section{Additional Link-Prediction Results}
\label{app:full-results}

We report the random-sampling results and the per-dataset new-node
comparison omitted from the main tables, followed by the full-test AP and AUC results with
standard deviations. All \method{} results use $k=8{,}192$, $R=4$, and three
runs; baseline results and their reported deviations are taken from the
published evaluations.

\subsection{Random Sampling and New-Node Interactions}

Table~\ref{tab:rq1-random} completes the full-test AP comparison under
random sampling. Table~\ref{tab:rq1-newnode} compares \method{} with the
best published result for each new-node dataset and regime.

\begin{table}[!htbp]
\caption{Full-test AP (\%) under random negative sampling. Values are
means; bold and underline mark the best and second-best displayed means,
and a dash indicates an unavailable published result.
Wins / 13 counts first places, including ties at the displayed precision.}
\label{tab:rq1-random}
\centering
\scriptsize
\setlength{\tabcolsep}{2.4pt}
\input{tables/rq1_random_wide}
\end{table}

\begin{table}[!htbp]
\caption{New-node AP (\%): the best published result and \method{} under
each regime. Values are means; $\Delta$ is \method{} minus the published
result in percentage points. Bold marks a \method{} lead;
Wins / 13 counts its first places, including ties at the displayed precision.}
\label{tab:rq1-newnode}
\centering
\scriptsize
\setlength{\tabcolsep}{3pt}
\input{tables/rq1_newnode_ap}
\end{table}

\subsection{Full-Test AP with Standard Deviations}

Tables~\ref{tab:full-historical}--\ref{tab:full-random} give the means and
standard deviations underlying the full-test AP comparisons. Each table
splits the thirteen datasets into two panels for readability.

\begin{table}[!htbp]
\caption{Full-test AP (\%) under historical negative sampling, reported as
mean $\pm$ standard deviation. Bold and underline mark the best and
second-best displayed means; a dash indicates an unavailable result.
\method{} uses three runs; baseline statistics are as published.}
\label{tab:full-historical}
\centering
\footnotesize
\setlength{\tabcolsep}{3pt}
\input{tables/rq1_historical_full}
\end{table}

\begin{table}[!htbp]
\caption{Full-test AP (\%) under inductive negative sampling, reported as
mean $\pm$ standard deviation. Conventions follow
Table~\ref{tab:full-historical}.}
\label{tab:full-inductive}
\centering
\footnotesize
\setlength{\tabcolsep}{3pt}
\input{tables/rq1_inductive_full}
\end{table}

\begin{table}[!htbp]
\caption{Full-test AP (\%) under random negative sampling, reported as
mean $\pm$ standard deviation. Conventions follow
Table~\ref{tab:full-historical}.}
\label{tab:full-random}
\centering
\footnotesize
\setlength{\tabcolsep}{3pt}
\input{tables/rq1_random_full}
\end{table}

\clearpage
\subsection{Full-Test AUC}
\label{app:auc}

Tables~\ref{tab:full-historical-auc}--\ref{tab:full-random-auc} report AUC
for the same full-test evaluation. \method{} exceeds the best available
published AUC on eleven datasets under historical sampling and ten under
inductive sampling. Under random sampling, it is within one AUC point of
the best published result on nine datasets. The AUC rankings differ from
AP on some datasets: for example, under historical sampling, \method{}
leads in AP on US Legis. but trails the best published AUC by 0.40 points.

\begin{table}[!htbp]
\caption{Full-test AUC (\%) under historical negative sampling, reported
as mean $\pm$ standard deviation. Conventions follow
Table~\ref{tab:full-historical}.}
\label{tab:full-historical-auc}
\centering
\footnotesize
\setlength{\tabcolsep}{3pt}
\input{tables/rq1_historical_auc_full}
\end{table}

\begin{table}[!htbp]
\caption{Full-test AUC (\%) under inductive negative sampling, reported
as mean $\pm$ standard deviation. Conventions follow
Table~\ref{tab:full-historical}.}
\label{tab:full-inductive-auc}
\centering
\footnotesize
\setlength{\tabcolsep}{3pt}
\input{tables/rq1_inductive_auc_full}
\end{table}

\begin{table}[!htbp]
\caption{Full-test AUC (\%) under random negative sampling, reported as
mean $\pm$ standard deviation. Conventions follow
Table~\ref{tab:full-historical}.}
\label{tab:full-random-auc}
\centering
\footnotesize
\setlength{\tabcolsep}{3pt}
\input{tables/rq1_random_auc_full}
\end{table}

\clearpage
\subsection{Predictor Comparison on All Datasets}
\label{app:predictors}

Table~\ref{tab:rq3-all} extends the predictor comparison of
Section~\ref{sec:rq3} to all thirteen datasets. All predictors receive the
same historical features and regime-matched context rows, with
$k=8{,}192$ and $R=4$. LightGBM and HGB are refitted at each refresh;
TabPFN and TabICL keep their pretrained parameters fixed. The additional
datasets include all four cases where LightGBM exceeds TabPFN: Can. Parl.
under random and historical sampling, and US Legis. and UN Vote under
historical sampling.

\begin{table}[!htbp]
\caption{AP (\%) of predictors given the same features and matched
context. Values are means over three runs; bold marks the best displayed
mean for each dataset and regime. TabPFN and LightGBM cover all thirteen
datasets; HGB and TabICL cover Wikipedia, Reddit, MOOC, LastFM, Enron, and
UCI. A dash indicates a predictor that was not evaluated on that dataset.}
\label{tab:rq3-all}
\centering
\scriptsize
\setlength{\tabcolsep}{3pt}
\input{tables/rq3_all}
\end{table}

\clearpage
\section{Context Refresh under Historical and Inductive Sampling}
\label{app:refresh}

\begin{figure}[H]
\centering
\includegraphics[width=\linewidth]{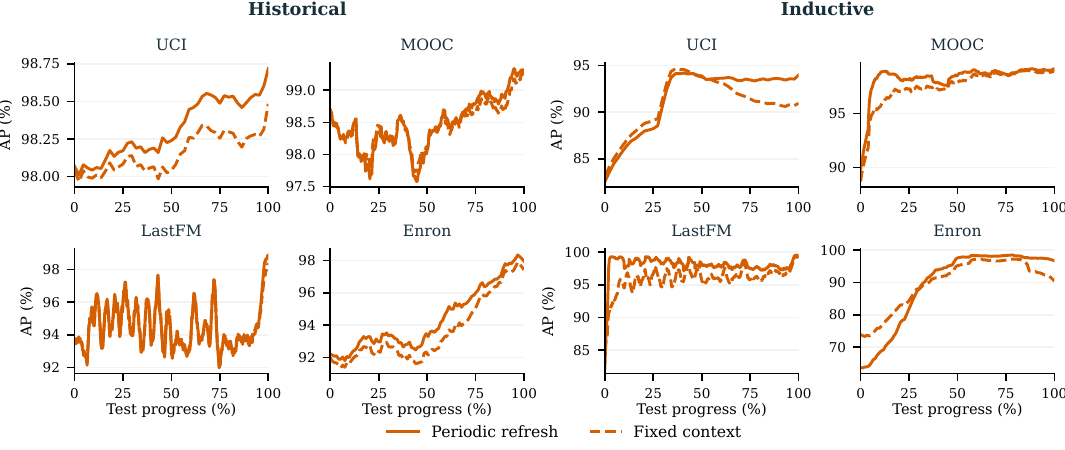}
\caption{AP over the test period under historical (left) and inductive
(right) negative sampling. Curves show mean AP over three runs with matched seeds, smoothed with
a centered 25-batch moving average and shorter windows at the boundaries.}
\label{fig:rq5-other}
\end{figure}

Figure~\ref{fig:rq5-other} supplements the random-sampling curves of
Figure~\ref{fig:rq5} with historical and inductive sampling on UCI, MOOC,
LastFM, and Enron. Both variants use $k=8{,}192$: periodic refresh uses
$R=4$, while the fixed variant keeps its initial context throughout the
test stream. Historical statistics and query features continue to update
in both variants.

Under historical sampling, the two curves follow similar fluctuations on
MOOC and nearly overlap on LastFM. The refreshed curve is slightly higher
on UCI and Enron, with differences that vary over the stream. Under
inductive sampling, the refreshed curve is higher through most of the
LastFM evaluation, while its early advantage on MOOC narrows later.
On UCI, the curves separate in the latter half as the fixed curve drops.
Enron shows a crossover: the fixed context starts ahead, but the refreshed
context is higher later. The near-zero overall gain on Enron thus combines differences of opposite
signs at different stages of evaluation.

\Needspace{370pt}
\section{Context Size and Refresh Period on Six Datasets}
\label{app:rq4-matrix}

\begin{figure}[H]
\centering
\includegraphics[width=\linewidth]{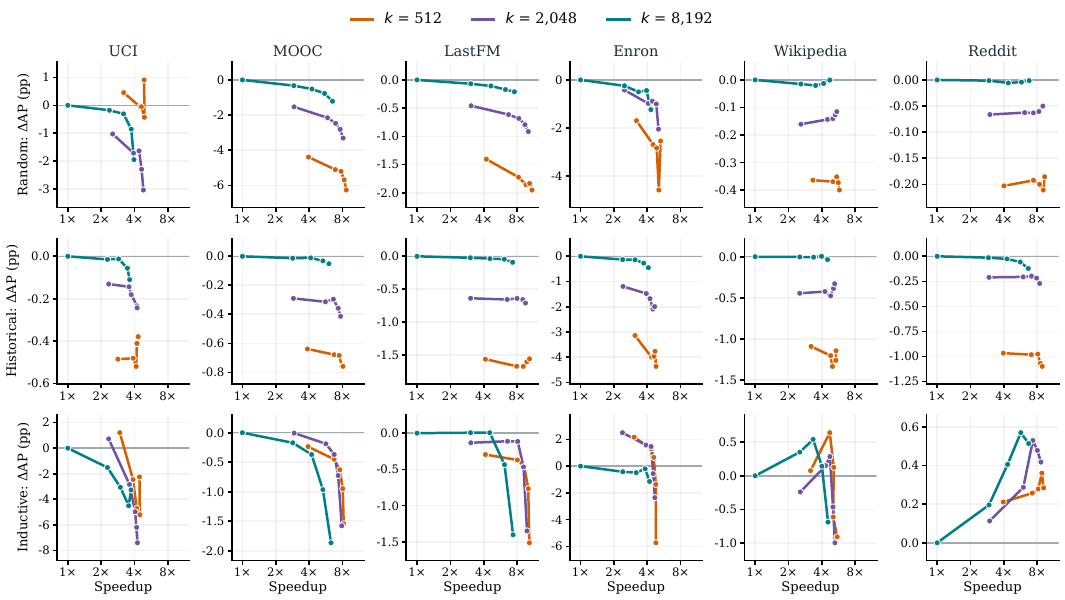}
\caption{Change in AP (percentage points) against evaluation speedup on
six datasets under random, historical, and inductive negative sampling.
Each line represents a context size $k$, with points for
$R=1,4,8,16,32$. AP differences and evaluation speedups are relative to
$k=8{,}192$, $R=1$ for the same dataset and sampling regime, using means
over three runs.}
\label{fig:rq4-all}
\end{figure}

Figure~\ref{fig:rq4-all} extends Figure~\ref{fig:rq4} to UCI, MOOC, LastFM,
Enron, Wikipedia, and Reddit. It shows all fifteen combinations of
$k\in\{512,2048,8192\}$ and $R\in\{1,4,8,16,32\}$ under each regime.
Accuracy and evaluation time are averaged over three runs for every
configuration; speedup is the ratio of mean full-stream evaluation times
relative to $k=8{,}192$, $R=1$ in the same dataset and regime.

The additional Wikipedia and Reddit panels show little change in AP along
the $k=8{,}192$ line under random and historical sampling, whereas smaller
contexts lose more AP under historical sampling. The inductive panels are
less uniform: some smaller contexts or longer refresh periods improve AP,
and increasing $R$ does not always reduce it. Together with the UCI and
Enron panels, these cases show why neither the largest context nor the
most frequent refresh is best in every setting.

%% file: tables/rq1_random_wide.tex
\begin{tabular}{@{}lcccccccccccccc@{}}
\toprule
Method & Wiki. & Reddit & MOOC & LastFM & Enron & S.Evo & UCI & Flights & CanP. & USLeg. & UNTr. & UNVo. & Cont. & Wins / 13 \\
\midrule
JODIE & 96.50 & 98.31 & 80.23 & 70.85 & 84.77 & 89.89 & 89.43 & 95.60 & 69.26 & 75.05 & 64.94 & 63.91 & 95.31 & 0 \\
DyRep & 94.86 & 98.22 & 81.97 & 71.92 & 82.38 & 88.87 & 65.14 & 95.29 & 66.54 & 75.34 & 63.21 & 62.81 & 95.98 & 0 \\
TGN & 98.45 & 98.63 & 89.15 & 77.07 & 86.53 & 93.57 & 92.34 & 97.95 & 70.88 & 75.99 & 65.03 & 65.72 & 96.89 & 0 \\
CTAN & 96.61 & 97.21 & 84.71 & 86.44 & 92.52 & --- & 76.64 & 93.64 & 87.01 & \underline{79.44} & 69.01 & 70.19 & 98.53 & 0 \\
CAWN & 98.76 & 99.11 & 80.15 & 86.99 & 89.56 & 84.96 & 95.18 & 98.51 & 69.82 & 70.58 & 65.39 & 52.84 & 90.26 & 0 \\
NAT & 98.03 & 99.13 & 85.88 & 88.02 & 90.60 & 88.92 & 93.40 & 98.57 & 79.72 & 78.71 & 73.95 & 70.45 & 97.39 & 0 \\
PINT & 98.45 & 99.15 & 88.08 & 89.66 & 92.20 & 94.42 & 96.45 & 98.80 & 68.36 & 74.85 & 70.20 & 66.25 & 98.64 & 0 \\
TPNet & \textbf{99.32} & 99.27 & \textbf{96.39} & \underline{94.50} & 92.90 & 94.73 & \underline{97.35} & 98.93 & 90.28 & \textbf{80.58} & \textbf{87.24} & \underline{75.12} & 98.66 & 4 \\
CNE-N & 99.09 & 99.22 & \underline{94.18} & 93.55 & 92.48 & 94.60 & 96.85 & 98.92 & 86.09 & 78.69 & 76.92 & 66.40 & 98.40 & 0 \\
TAWRMAC & 99.29 & 99.30 & 91.46 & 91.17 & 92.74 & --- & 96.40 & 98.84 & 76.42 & 73.89 & 68.11 & 66.50 & \textbf{98.80} & 1 \\
TGAT & 96.94 & 98.52 & 85.84 & 73.42 & 71.12 & 93.16 & 79.63 & 94.03 & 70.73 & 68.52 & 61.47 & 52.21 & 96.28 & 0 \\
TCL & 96.47 & 97.53 & 82.38 & 67.27 & 79.70 & 93.13 & 89.57 & 91.23 & 68.67 & 69.59 & 62.21 & 51.90 & 92.44 & 0 \\
GraphMixer & 97.25 & 97.31 & 82.78 & 75.61 & 82.25 & 93.37 & 93.25 & 90.99 & 77.04 & 70.74 & 62.61 & 52.11 & 91.92 & 0 \\
DyGFormer & 99.03 & 99.22 & 87.52 & 93.00 & 92.47 & 94.73 & 95.79 & 98.91 & 97.36 & 71.11 & 66.46 & 55.55 & 98.29 & 0 \\
FreeDyG & 99.26 & \textbf{99.48} & 89.61 & 92.15 & 92.51 & \textbf{94.91} & 96.28 & --- & --- & --- & --- & --- & --- & 2 \\
DyGMamba & 99.15 & 99.32 & 89.21 & 93.35 & 92.65 & \underline{94.77} & 95.91 & \textbf{98.95} & \textbf{99.57} & 71.75 & 67.50 & 56.39 & 98.43 & 2 \\
RepeatMixer & 99.16 & 99.22 & 92.76 & 94.14 & 92.66 & --- & 96.74 & --- & --- & --- & --- & --- & --- & 0 \\
TIDFormer & \underline{99.30} & \underline{99.38} & 93.42 & \textbf{94.60} & \underline{93.17} & 94.76 & \textbf{97.44} & --- & --- & --- & --- & --- & --- & 2 \\
TAMI-GraphMixer & 98.89 & 98.84 & 83.53 & 88.13 & 90.97 & 93.41 & 96.20 & 96.75 & 78.38 & 71.57 & 62.98 & 57.74 & 95.26 & 0 \\
TAMI-DyGFormer & 99.25 & 99.29 & 88.49 & 94.03 & 92.66 & 94.74 & 96.72 & \underline{98.94} & \underline{98.67} & 71.40 & 66.39 & 56.02 & 98.70 & 0 \\
EdgeBank & 90.37 & 94.86 & 57.97 & 79.29 & 83.53 & 74.95 & 76.20 & 89.35 & 64.55 & 58.39 & 60.41 & 58.49 & 92.58 & 0 \\
\midrule
\method{} & 98.63 & 99.16 & 83.68 & 92.37 & \textbf{93.83} & 94.11 & 91.44 & 98.55 & 84.33 & 78.94 & \underline{85.96} & \textbf{77.74} & \underline{98.79} & 2 \\
\bottomrule
\end{tabular}

%% file: tables/rq1_newnode_ap.tex
\begin{tabular}{@{}l ll r ll r ll r@{}}
\toprule
 & \multicolumn{3}{c}{Historical} & \multicolumn{3}{c}{Inductive} & \multicolumn{3}{c}{Random} \\
\cmidrule(lr){2-4}\cmidrule(lr){5-7}\cmidrule(lr){8-10}
Dataset & Best baseline & \method{} & $\Delta$ & Best baseline & \method{} & $\Delta$ & Best baseline & \method{} & $\Delta$ \\
\midrule
Wikipedia & GraphMixer 87.60 & \textbf{90.59} & +2.99 & CTAN 92.61 & 75.98 & $-$16.63 & FreeDyG 98.97 & 97.70 & $-$1.27 \\
Reddit & TAWRMAC 67.95 & \textbf{83.57} & +15.62 & CTAN 78.35 & 65.38 & $-$12.97 & TAWRMAC 98.99 & 98.74 & $-$0.25 \\
MOOC & TIDFormer 84.61 & \textbf{95.21} & +10.60 & TIDFormer 84.61 & \textbf{90.91} & +6.30 & TPNet 95.07 & 83.56 & $-$11.51 \\
LastFM & TIDFormer 82.79 & \textbf{88.94} & +6.15 & TIDFormer 82.79 & 54.83 & $-$27.96 & TIDFormer 95.41 & 93.61 & $-$1.80 \\
Enron & RepeatMixer 84.49 & 74.17 & $-$10.32 & RepeatMixer 84.48 & 68.47 & $-$16.01 & TPNet 90.34 & 89.85 & $-$0.49 \\
Social Evo. & DyGFormer 96.82 & \textbf{98.83} & +2.01 & FreeDyG 96.91 & 94.06 & $-$2.85 & FreeDyG 94.76 & 91.55 & $-$3.21 \\
UCI & TIDFormer 86.11 & \textbf{89.53} & +3.42 & TIDFormer 86.11 & 81.15 & $-$4.96 & TIDFormer 95.95 & 82.33 & $-$13.62 \\
Flights & GraphMixer 65.28 & \textbf{87.57} & +22.29 & CTAN 66.53 & \textbf{83.58} & +17.05 & TPNet 97.97 & 97.08 & $-$0.89 \\
Can. Parl. & DyGFormer 87.40 & 68.11 & $-$19.29 & DyGMamba 92.68 & 70.87 & $-$21.81 & DyGMamba 93.46 & 64.73 & $-$28.73 \\
US Legis. & TPNet 66.95 & 65.54 & $-$1.41 & TPNet 66.95 & 63.61 & $-$3.34 & TPNet 61.71 & \textbf{63.26} & +1.55 \\
UN Trade & TPNet 78.83 & 75.79 & $-$3.04 & TPNet 78.95 & \textbf{84.13} & +5.18 & TPNet 86.53 & 86.08 & $-$0.45 \\
UN Vote & PINT 65.75 & \textbf{84.10} & +18.35 & TAWRMAC 66.71 & \textbf{86.51} & +19.80 & CNE-N 69.71 & \textbf{77.87} & +8.16 \\
Contact & TGAT 94.15 & \textbf{98.30} & +4.15 & TGAT 94.14 & 92.02 & $-$2.12 & TPNet 98.39 & \textbf{98.54} & +0.15 \\
\midrule
Wins / 13 & & 9 & & & 4 & & & 3 & \\
\bottomrule
\end{tabular}

%% file: tables/rq1_historical_full.tex
\begin{tabular}{@{}lccccccc@{}}
\toprule
Method & Wikipedia & Reddit & MOOC & LastFM & Enron & Social Evo. & UCI \\
\midrule
JODIE & 83.01 $\pm$ 0.66 & 80.03 $\pm$ 0.36 & 78.94 $\pm$ 1.25 & 74.35 $\pm$ 3.81 & 69.85 $\pm$ 2.70 & 87.44 $\pm$ 6.78 & 75.24 $\pm$ 5.80 \\
DyRep & 79.93 $\pm$ 0.56 & 79.83 $\pm$ 0.31 & 75.60 $\pm$ 1.12 & 74.92 $\pm$ 2.46 & 71.19 $\pm$ 2.76 & 93.29 $\pm$ 0.43 & 55.10 $\pm$ 3.14 \\
TGN & 86.86 $\pm$ 0.33 & 81.22 $\pm$ 0.61 & 87.06 $\pm$ 1.93 & 76.87 $\pm$ 4.64 & 73.91 $\pm$ 1.76 & 94.45 $\pm$ 0.56 & 80.43 $\pm$ 2.12 \\
CTAN & \underline{95.91 $\pm$ 0.10} & \underline{89.77 $\pm$ 2.28} & 67.73 $\pm$ 2.08 & 82.29 $\pm$ 0.94 & 77.24 $\pm$ 1.53 & --- & 76.62 $\pm$ 0.33 \\
CAWN & 71.21 $\pm$ 1.67 & 80.82 $\pm$ 0.45 & 74.05 $\pm$ 0.95 & 69.86 $\pm$ 0.43 & 64.73 $\pm$ 0.36 & 85.53 $\pm$ 0.38 & 65.30 $\pm$ 0.43 \\
NAT & 68.36 $\pm$ 4.47 & 80.58 $\pm$ 0.91 & 85.42 $\pm$ 3.26 & 78.75 $\pm$ 1.63 & 72.90 $\pm$ 1.38 & 91.29 $\pm$ 4.89 & 75.38 $\pm$ 1.28 \\
PINT & 91.33 $\pm$ 0.76 & 83.93 $\pm$ 1.11 & 84.80 $\pm$ 1.57 & 84.95 $\pm$ 0.43 & 85.41 $\pm$ 0.84 & 96.73 $\pm$ 0.29 & \underline{93.96 $\pm$ 0.20} \\
TPNet & 81.55 $\pm$ 4.10 & 81.02 $\pm$ 1.31 & 92.69 $\pm$ 0.95 & 87.74 $\pm$ 0.50 & 80.79 $\pm$ 1.68 & 96.80 $\pm$ 0.41 & 86.34 $\pm$ 0.80 \\
CNE-N & --- & --- & --- & --- & --- & --- & --- \\
TAWRMAC & 90.16 $\pm$ 0.27 & 84.36 $\pm$ 0.43 & 88.93 $\pm$ 0.39 & 82.69 $\pm$ 0.56 & 80.23 $\pm$ 0.79 & --- & 88.29 $\pm$ 0.14 \\
TGAT & 87.38 $\pm$ 0.22 & 79.55 $\pm$ 0.20 & 82.19 $\pm$ 0.62 & 71.59 $\pm$ 0.24 & 64.07 $\pm$ 1.05 & 95.01 $\pm$ 0.44 & 68.27 $\pm$ 1.37 \\
TCL & 89.05 $\pm$ 0.39 & 77.14 $\pm$ 0.16 & 77.06 $\pm$ 0.41 & 59.30 $\pm$ 2.31 & 70.66 $\pm$ 0.39 & 94.74 $\pm$ 0.31 & 80.25 $\pm$ 2.74 \\
GraphMixer & 90.90 $\pm$ 0.10 & 78.44 $\pm$ 0.18 & 77.77 $\pm$ 0.92 & 72.47 $\pm$ 0.49 & 77.98 $\pm$ 0.92 & 94.93 $\pm$ 0.31 & 84.11 $\pm$ 1.35 \\
DyGFormer & 82.23 $\pm$ 2.54 & 81.57 $\pm$ 0.67 & 85.85 $\pm$ 0.66 & 81.57 $\pm$ 0.48 & 75.63 $\pm$ 0.73 & 97.38 $\pm$ 0.14 & 82.17 $\pm$ 0.82 \\
FreeDyG & 91.59 $\pm$ 0.57 & 85.67 $\pm$ 1.01 & 86.71 $\pm$ 0.81 & 79.71 $\pm$ 0.51 & 78.87 $\pm$ 0.82 & \underline{97.79 $\pm$ 0.23} & 86.10 $\pm$ 1.19 \\
DyGMamba & 81.77 $\pm$ 1.20 & 81.80 $\pm$ 1.52 & 85.89 $\pm$ 0.94 & 83.02 $\pm$ 0.16 & 77.77 $\pm$ 1.32 & 97.35 $\pm$ 0.52 & 81.03 $\pm$ 1.09 \\
RepeatMixer & 91.02 $\pm$ 0.59 & 84.44 $\pm$ 0.50 & 94.24 $\pm$ 0.25 & 88.41 $\pm$ 0.07 & \underline{88.19 $\pm$ 0.23} & --- & 87.23 $\pm$ 0.23 \\
TIDFormer & 91.21 $\pm$ 0.78 & 83.27 $\pm$ 1.01 & \underline{95.37 $\pm$ 1.09} & \underline{88.68 $\pm$ 0.57} & 81.59 $\pm$ 1.17 & 97.39 $\pm$ 0.67 & 89.57 $\pm$ 1.17 \\
TAMI-GraphMixer & 90.97 $\pm$ 0.57 & 82.56 $\pm$ 0.29 & 83.61 $\pm$ 0.62 & 80.23 $\pm$ 1.06 & 81.68 $\pm$ 0.68 & 96.80 $\pm$ 0.03 & 87.69 $\pm$ 0.53 \\
TAMI-DyGFormer & 82.38 $\pm$ 0.24 & 85.18 $\pm$ 0.10 & 86.26 $\pm$ 0.89 & 83.40 $\pm$ 1.84 & 81.02 $\pm$ 0.01 & 97.56 $\pm$ 0.04 & 85.89 $\pm$ 0.32 \\
EdgeBank & 73.35 $\pm$ 0.00 & 73.59 $\pm$ 0.00 & 60.71 $\pm$ 0.00 & 73.03 $\pm$ 0.00 & 76.53 $\pm$ 0.00 & 80.57 $\pm$ 0.00 & 65.50 $\pm$ 0.00 \\
\midrule
\method{} & \textbf{96.65 $\pm$ 0.02} & \textbf{93.78 $\pm$ 0.02} & \textbf{98.48 $\pm$ 0.00} & \textbf{94.48 $\pm$ 0.01} & \textbf{94.40 $\pm$ 0.08} & \textbf{99.50 $\pm$ 0.00} & \textbf{98.34 $\pm$ 0.01} \\
\bottomrule
\end{tabular}
\par\medskip
\begin{tabular}{@{}lcccccc@{}}
\toprule
Method & Flights & Can. Parl. & US Legis. & UN Trade & UN Vote & Contact \\
\midrule
JODIE & 66.48 $\pm$ 2.59 & 51.79 $\pm$ 0.63 & 51.71 $\pm$ 5.76 & 61.39 $\pm$ 1.83 & 70.02 $\pm$ 0.81 & 95.31 $\pm$ 2.13 \\
DyRep & 67.61 $\pm$ 0.99 & 63.31 $\pm$ 1.23 & 86.88 $\pm$ 2.25 & 59.19 $\pm$ 1.07 & 69.30 $\pm$ 1.12 & 96.39 $\pm$ 0.20 \\
TGN & 66.70 $\pm$ 1.64 & 68.42 $\pm$ 3.07 & 74.00 $\pm$ 7.57 & 58.44 $\pm$ 5.51 & 69.37 $\pm$ 3.93 & 93.05 $\pm$ 2.35 \\
CTAN & \underline{75.75 $\pm$ 8.04} & 89.28 $\pm$ 1.24 & 84.20 $\pm$ 0.95 & 67.38 $\pm$ 1.08 & 74.75 $\pm$ 1.81 & 97.86 $\pm$ 0.15 \\
CAWN & 64.72 $\pm$ 0.97 & 66.53 $\pm$ 2.77 & 68.82 $\pm$ 8.23 & 55.71 $\pm$ 0.38 & 51.26 $\pm$ 0.04 & 84.16 $\pm$ 0.49 \\
NAT & 64.74 $\pm$ 1.83 & 77.72 $\pm$ 1.78 & 91.12 $\pm$ 1.97 & 78.65 $\pm$ 1.16 & 71.39 $\pm$ 2.68 & 96.84 $\pm$ 0.57 \\
PINT & 66.82 $\pm$ 1.44 & 63.84 $\pm$ 5.36 & 73.58 $\pm$ 4.16 & 70.07 $\pm$ 2.28 & 71.79 $\pm$ 2.53 & 97.61 $\pm$ 0.18 \\
TPNet & 69.10 $\pm$ 1.27 & 86.61 $\pm$ 3.57 & \underline{94.55 $\pm$ 0.62} & \underline{85.22 $\pm$ 1.22} & 74.68 $\pm$ 1.38 & \underline{98.02 $\pm$ 0.15} \\
CNE-N & --- & --- & --- & --- & --- & --- \\
TAWRMAC & 68.57 $\pm$ 0.56 & 76.60 $\pm$ 2.38 & 91.17 $\pm$ 2.58 & 67.99 $\pm$ 2.68 & 74.67 $\pm$ 2.58 & 97.81 $\pm$ 0.09 \\
TGAT & 72.38 $\pm$ 0.18 & 67.13 $\pm$ 0.84 & 62.14 $\pm$ 6.60 & 55.74 $\pm$ 0.91 & 52.96 $\pm$ 2.14 & 96.05 $\pm$ 0.52 \\
TCL & 70.68 $\pm$ 0.24 & 65.93 $\pm$ 3.00 & 80.53 $\pm$ 3.95 & 55.90 $\pm$ 1.17 & 52.30 $\pm$ 2.35 & 93.86 $\pm$ 0.21 \\
GraphMixer & 71.47 $\pm$ 0.26 & 74.34 $\pm$ 0.87 & 81.65 $\pm$ 1.02 & 57.05 $\pm$ 1.22 & 51.20 $\pm$ 1.60 & 93.36 $\pm$ 0.41 \\
DyGFormer & 66.59 $\pm$ 0.49 & 97.00 $\pm$ 0.31 & 85.30 $\pm$ 3.88 & 64.41 $\pm$ 1.40 & 60.84 $\pm$ 1.58 & 97.57 $\pm$ 0.06 \\
FreeDyG & --- & --- & --- & --- & --- & --- \\
DyGMamba & 67.80 $\pm$ 2.17 & \textbf{99.77 $\pm$ 0.12} & 82.15 $\pm$ 1.02 & 65.10 $\pm$ 0.02 & 61.07 $\pm$ 0.39 & 97.61 $\pm$ 0.04 \\
RepeatMixer & --- & --- & --- & --- & --- & --- \\
TIDFormer & --- & --- & --- & --- & --- & --- \\
TAMI-GraphMixer & 73.01 $\pm$ 0.11 & 78.81 $\pm$ 0.12 & 84.56 $\pm$ 1.11 & 69.74 $\pm$ 2.15 & 70.90 $\pm$ 0.32 & 93.30 $\pm$ 0.02 \\
TAMI-DyGFormer & 67.77 $\pm$ 2.93 & \underline{98.96 $\pm$ 0.01} & 86.10 $\pm$ 2.50 & 65.16 $\pm$ 1.46 & 81.72 $\pm$ 2.12 & 97.72 $\pm$ 0.02 \\
EdgeBank & 70.53 $\pm$ 0.00 & 63.84 $\pm$ 0.00 & 63.22 $\pm$ 0.00 & 81.32 $\pm$ 0.00 & \underline{84.89 $\pm$ 0.00} & 88.81 $\pm$ 0.00 \\
\midrule
\method{} & \textbf{94.96 $\pm$ 0.31} & 98.02 $\pm$ 0.09 & \textbf{95.26 $\pm$ 0.29} & \textbf{91.34 $\pm$ 0.16} & \textbf{90.46 $\pm$ 0.25} & \textbf{99.46 $\pm$ 0.00} \\
\bottomrule
\end{tabular}

%% file: tables/rq1_inductive_full.tex
\begin{tabular}{@{}lccccccc@{}}
\toprule
Method & Wikipedia & Reddit & MOOC & LastFM & Enron & Social Evo. & UCI \\
\midrule
JODIE & 75.65 $\pm$ 0.79 & 86.98 $\pm$ 0.16 & 65.23 $\pm$ 2.19 & 62.67 $\pm$ 4.49 & 68.96 $\pm$ 0.98 & 89.82 $\pm$ 4.11 & 65.99 $\pm$ 1.40 \\
DyRep & 70.21 $\pm$ 1.58 & 86.30 $\pm$ 0.26 & 61.66 $\pm$ 0.95 & 64.41 $\pm$ 2.70 & 67.79 $\pm$ 1.53 & 93.28 $\pm$ 0.48 & 54.79 $\pm$ 1.76 \\
TGN & 85.62 $\pm$ 0.44 & 88.10 $\pm$ 0.24 & 77.50 $\pm$ 2.91 & 65.95 $\pm$ 5.98 & 70.89 $\pm$ 2.72 & 95.13 $\pm$ 0.56 & 70.94 $\pm$ 0.71 \\
CTAN & \underline{94.15 $\pm$ 0.08} & 90.99 $\pm$ 2.19 & 64.93 $\pm$ 3.31 & 80.06 $\pm$ 0.85 & 72.02 $\pm$ 2.64 & --- & 66.25 $\pm$ 0.51 \\
CAWN & 74.06 $\pm$ 2.62 & 91.67 $\pm$ 0.24 & 73.51 $\pm$ 0.94 & 67.48 $\pm$ 0.77 & 75.15 $\pm$ 0.58 & 88.32 $\pm$ 0.27 & 64.61 $\pm$ 0.48 \\
NAT & 53.85 $\pm$ 6.25 & 81.71 $\pm$ 1.03 & 77.28 $\pm$ 1.94 & 64.02 $\pm$ 1.48 & 70.34 $\pm$ 1.73 & 92.12 $\pm$ 4.53 & 57.55 $\pm$ 0.81 \\
PINT & 82.39 $\pm$ 2.70 & 87.59 $\pm$ 0.65 & 75.70 $\pm$ 1.43 & 72.24 $\pm$ 0.92 & 80.23 $\pm$ 1.37 & 97.15 $\pm$ 0.25 & 83.61 $\pm$ 0.23 \\
TPNet & 79.35 $\pm$ 5.52 & 88.19 $\pm$ 0.33 & 88.18 $\pm$ 0.97 & 77.99 $\pm$ 1.30 & 75.36 $\pm$ 1.81 & 97.35 $\pm$ 0.32 & 77.26 $\pm$ 1.57 \\
CNE-N & --- & --- & --- & --- & --- & --- & --- \\
TAWRMAC & 89.55 $\pm$ 1.37 & 90.80 $\pm$ 0.28 & 83.42 $\pm$ 2.57 & 72.57 $\pm$ 0.37 & 78.58 $\pm$ 0.59 & --- & 81.46 $\pm$ 0.90 \\
TGAT & 87.00 $\pm$ 0.16 & 89.59 $\pm$ 0.24 & 75.95 $\pm$ 0.64 & 71.13 $\pm$ 0.17 & 63.94 $\pm$ 1.36 & 94.84 $\pm$ 0.44 & 68.67 $\pm$ 0.84 \\
TCL & 86.76 $\pm$ 0.72 & 87.45 $\pm$ 0.29 & 74.65 $\pm$ 0.54 & 58.21 $\pm$ 0.89 & 71.29 $\pm$ 0.32 & 94.90 $\pm$ 0.36 & 76.01 $\pm$ 1.11 \\
GraphMixer & 88.59 $\pm$ 0.17 & 85.26 $\pm$ 0.11 & 74.27 $\pm$ 0.92 & 68.12 $\pm$ 0.33 & 75.01 $\pm$ 0.79 & 94.72 $\pm$ 0.33 & 80.10 $\pm$ 0.51 \\
DyGFormer & 78.29 $\pm$ 5.38 & 91.11 $\pm$ 0.40 & 81.24 $\pm$ 0.69 & 73.97 $\pm$ 0.50 & 77.41 $\pm$ 0.89 & 97.68 $\pm$ 0.10 & 72.25 $\pm$ 1.71 \\
FreeDyG & 90.05 $\pm$ 0.79 & 90.74 $\pm$ 0.17 & 83.01 $\pm$ 0.87 & 72.19 $\pm$ 0.24 & 77.81 $\pm$ 0.65 & 97.57 $\pm$ 0.15 & 82.35 $\pm$ 0.73 \\
DyGMamba & 79.86 $\pm$ 2.18 & 91.15 $\pm$ 0.54 & 81.11 $\pm$ 0.63 & 73.63 $\pm$ 0.54 & 80.86 $\pm$ 1.24 & 97.68 $\pm$ 0.42 & 71.95 $\pm$ 2.51 \\
RepeatMixer & 88.98 $\pm$ 1.49 & 91.76 $\pm$ 0.56 & 83.68 $\pm$ 0.72 & 76.34 $\pm$ 0.30 & 84.17 $\pm$ 0.24 & --- & 84.96 $\pm$ 0.52 \\
TIDFormer & 91.15 $\pm$ 1.17 & 91.57 $\pm$ 0.50 & 84.53 $\pm$ 1.11 & 75.69 $\pm$ 1.17 & 79.67 $\pm$ 0.66 & 97.70 $\pm$ 0.36 & \underline{85.81 $\pm$ 1.39} \\
TAMI-GraphMixer & 93.89 $\pm$ 0.24 & \underline{96.19 $\pm$ 0.09} & 79.82 $\pm$ 0.38 & \underline{91.06 $\pm$ 0.06} & \underline{88.95 $\pm$ 0.29} & 96.09 $\pm$ 0.00 & 84.12 $\pm$ 0.39 \\
TAMI-DyGFormer & 78.96 $\pm$ 0.08 & 94.37 $\pm$ 0.21 & \underline{92.39 $\pm$ 1.38} & 74.03 $\pm$ 1.63 & 86.23 $\pm$ 1.39 & \underline{97.76 $\pm$ 0.21} & 80.13 $\pm$ 0.35 \\
EdgeBank & 80.63 $\pm$ 0.00 & 85.48 $\pm$ 0.00 & 49.43 $\pm$ 0.00 & 75.49 $\pm$ 0.00 & 73.89 $\pm$ 0.00 & 83.69 $\pm$ 0.00 & 57.43 $\pm$ 0.00 \\
\midrule
\method{} & \textbf{97.39 $\pm$ 0.07} & \textbf{99.13 $\pm$ 0.05} & \textbf{98.21 $\pm$ 0.04} & \textbf{98.00 $\pm$ 0.01} & \textbf{89.49 $\pm$ 0.73} & \textbf{97.81 $\pm$ 0.33} & \textbf{90.73 $\pm$ 0.08} \\
\bottomrule
\end{tabular}
\par\medskip
\begin{tabular}{@{}lcccccc@{}}
\toprule
Method & Flights & Can. Parl. & US Legis. & UN Trade & UN Vote & Contact \\
\midrule
JODIE & 69.07 $\pm$ 4.02 & 48.42 $\pm$ 0.66 & 50.27 $\pm$ 5.13 & 60.42 $\pm$ 1.48 & 67.79 $\pm$ 1.46 & 93.43 $\pm$ 1.78 \\
DyRep & 70.57 $\pm$ 1.82 & 58.61 $\pm$ 0.86 & 83.44 $\pm$ 1.16 & 60.19 $\pm$ 1.24 & 67.53 $\pm$ 1.98 & 94.18 $\pm$ 0.10 \\
TGN & 71.09 $\pm$ 2.72 & 65.34 $\pm$ 2.87 & 67.57 $\pm$ 6.47 & 61.04 $\pm$ 6.01 & 67.63 $\pm$ 2.67 & 90.18 $\pm$ 3.28 \\
CTAN & 76.20 $\pm$ 7.96 & 87.42 $\pm$ 0.71 & 79.13 $\pm$ 2.03 & 68.98 $\pm$ 0.73 & 72.85 $\pm$ 1.36 & 97.31 $\pm$ 0.19 \\
CAWN & 69.18 $\pm$ 1.52 & 67.75 $\pm$ 1.00 & 65.81 $\pm$ 8.52 & 62.54 $\pm$ 0.67 & 52.19 $\pm$ 0.34 & 89.31 $\pm$ 0.27 \\
NAT & 59.16 $\pm$ 1.48 & 78.03 $\pm$ 1.27 & \underline{88.40 $\pm$ 2.34} & 78.38 $\pm$ 2.24 & 72.53 $\pm$ 1.94 & 91.13 $\pm$ 1.48 \\
PINT & 61.99 $\pm$ 1.70 & 59.15 $\pm$ 6.20 & 67.78 $\pm$ 4.29 & 69.72 $\pm$ 1.84 & 68.37 $\pm$ 1.92 & 94.05 $\pm$ 0.52 \\
TPNet & 64.78 $\pm$ 1.50 & 85.59 $\pm$ 3.08 & \textbf{91.05 $\pm$ 1.21} & 86.61 $\pm$ 0.99 & 75.05 $\pm$ 1.41 & 95.84 $\pm$ 0.32 \\
CNE-N & --- & --- & --- & --- & --- & --- \\
TAWRMAC & 65.62 $\pm$ 1.25 & 69.87 $\pm$ 1.80 & 84.02 $\pm$ 1.98 & 60.64 $\pm$ 1.26 & 69.63 $\pm$ 1.98 & 96.01 $\pm$ 0.10 \\
TGAT & 75.48 $\pm$ 0.26 & 68.82 $\pm$ 1.21 & 61.91 $\pm$ 5.82 & 60.61 $\pm$ 1.24 & 52.89 $\pm$ 1.61 & 94.35 $\pm$ 0.48 \\
TCL & 74.62 $\pm$ 0.18 & 65.85 $\pm$ 1.75 & 78.15 $\pm$ 3.34 & 61.06 $\pm$ 1.74 & 50.62 $\pm$ 0.82 & 91.35 $\pm$ 0.21 \\
GraphMixer & 74.87 $\pm$ 0.21 & 69.48 $\pm$ 0.63 & 79.63 $\pm$ 0.84 & 60.15 $\pm$ 1.29 & 51.60 $\pm$ 0.73 & 90.87 $\pm$ 0.35 \\
DyGFormer & 70.92 $\pm$ 1.78 & 95.44 $\pm$ 0.57 & 81.25 $\pm$ 3.62 & 55.79 $\pm$ 1.02 & 51.91 $\pm$ 0.84 & 94.75 $\pm$ 0.28 \\
FreeDyG & --- & --- & --- & --- & --- & --- \\
DyGMamba & 73.79 $\pm$ 5.69 & \textbf{98.32 $\pm$ 0.34} & 81.67 $\pm$ 2.16 & 58.89 $\pm$ 0.98 & 52.24 $\pm$ 0.95 & 95.43 $\pm$ 0.17 \\
RepeatMixer & --- & --- & --- & --- & --- & --- \\
TIDFormer & --- & --- & --- & --- & --- & --- \\
TAMI-GraphMixer & \underline{93.64 $\pm$ 0.69} & 70.94 $\pm$ 0.28 & 83.31 $\pm$ 0.01 & \underline{87.73 $\pm$ 0.68} & \underline{79.53 $\pm$ 0.04} & 96.12 $\pm$ 0.74 \\
TAMI-DyGFormer & 75.55 $\pm$ 2.82 & \underline{97.25 $\pm$ 0.09} & 81.31 $\pm$ 2.86 & 68.01 $\pm$ 3.27 & 78.19 $\pm$ 1.44 & \underline{98.47 $\pm$ 0.06} \\
EdgeBank & 81.08 $\pm$ 0.00 & 62.16 $\pm$ 0.00 & 64.74 $\pm$ 0.00 & 72.97 $\pm$ 0.00 & 66.30 $\pm$ 0.00 & 85.20 $\pm$ 0.00 \\
\midrule
\method{} & \textbf{98.87 $\pm$ 0.29} & 83.16 $\pm$ 0.19 & 83.09 $\pm$ 0.32 & \textbf{95.33 $\pm$ 0.22} & \textbf{92.38 $\pm$ 0.05} & \textbf{99.79 $\pm$ 0.00} \\
\bottomrule
\end{tabular}

%% file: tables/rq1_random_full.tex
\begin{tabular}{@{}lccccccc@{}}
\toprule
Method & Wikipedia & Reddit & MOOC & LastFM & Enron & Social Evo. & UCI \\
\midrule
JODIE & 96.50 $\pm$ 0.14 & 98.31 $\pm$ 0.14 & 80.23 $\pm$ 2.44 & 70.85 $\pm$ 2.13 & 84.77 $\pm$ 0.30 & 89.89 $\pm$ 0.55 & 89.43 $\pm$ 1.09 \\
DyRep & 94.86 $\pm$ 0.06 & 98.22 $\pm$ 0.04 & 81.97 $\pm$ 0.49 & 71.92 $\pm$ 2.21 & 82.38 $\pm$ 3.36 & 88.87 $\pm$ 0.30 & 65.14 $\pm$ 2.30 \\
TGN & 98.45 $\pm$ 0.06 & 98.63 $\pm$ 0.06 & 89.15 $\pm$ 1.60 & 77.07 $\pm$ 3.97 & 86.53 $\pm$ 1.11 & 93.57 $\pm$ 0.17 & 92.34 $\pm$ 1.04 \\
CTAN & 96.61 $\pm$ 0.79 & 97.21 $\pm$ 0.84 & 84.71 $\pm$ 2.85 & 86.44 $\pm$ 0.80 & 92.52 $\pm$ 1.20 & --- & 76.64 $\pm$ 4.11 \\
CAWN & 98.76 $\pm$ 0.03 & 99.11 $\pm$ 0.01 & 80.15 $\pm$ 0.25 & 86.99 $\pm$ 0.06 & 89.56 $\pm$ 0.09 & 84.96 $\pm$ 0.09 & 95.18 $\pm$ 0.06 \\
NAT & 98.03 $\pm$ 0.07 & 99.13 $\pm$ 0.10 & 85.88 $\pm$ 0.55 & 88.02 $\pm$ 1.94 & 90.60 $\pm$ 0.66 & 88.92 $\pm$ 3.45 & 93.40 $\pm$ 0.26 \\
PINT & 98.45 $\pm$ 0.04 & 99.15 $\pm$ 0.02 & 88.08 $\pm$ 0.86 & 89.66 $\pm$ 1.81 & 92.20 $\pm$ 0.15 & 94.42 $\pm$ 0.03 & 96.45 $\pm$ 0.11 \\
TPNet & \textbf{99.32 $\pm$ 0.03} & 99.27 $\pm$ 0.00 & \textbf{96.39 $\pm$ 0.09} & \underline{94.50 $\pm$ 0.08} & 92.90 $\pm$ 0.17 & 94.73 $\pm$ 0.02 & \underline{97.35 $\pm$ 0.04} \\
CNE-N & 99.09 $\pm$ 0.04 & 99.22 $\pm$ 0.01 & \underline{94.18 $\pm$ 0.07} & 93.55 $\pm$ 0.12 & 92.48 $\pm$ 0.10 & 94.60 $\pm$ 0.03 & 96.85 $\pm$ 0.08 \\
TAWRMAC & 99.29 $\pm$ 0.02 & 99.30 $\pm$ 0.01 & 91.46 $\pm$ 0.66 & 91.17 $\pm$ 0.08 & 92.74 $\pm$ 0.21 & --- & 96.40 $\pm$ 0.15 \\
TGAT & 96.94 $\pm$ 0.06 & 98.52 $\pm$ 0.02 & 85.84 $\pm$ 0.15 & 73.42 $\pm$ 0.21 & 71.12 $\pm$ 0.97 & 93.16 $\pm$ 0.17 & 79.63 $\pm$ 0.70 \\
TCL & 96.47 $\pm$ 0.16 & 97.53 $\pm$ 0.02 & 82.38 $\pm$ 0.24 & 67.27 $\pm$ 2.16 & 79.70 $\pm$ 0.71 & 93.13 $\pm$ 0.16 & 89.57 $\pm$ 1.63 \\
GraphMixer & 97.25 $\pm$ 0.03 & 97.31 $\pm$ 0.01 & 82.78 $\pm$ 0.15 & 75.61 $\pm$ 0.24 & 82.25 $\pm$ 0.16 & 93.37 $\pm$ 0.07 & 93.25 $\pm$ 0.57 \\
DyGFormer & 99.03 $\pm$ 0.02 & 99.22 $\pm$ 0.01 & 87.52 $\pm$ 0.49 & 93.00 $\pm$ 0.12 & 92.47 $\pm$ 0.12 & 94.73 $\pm$ 0.01 & 95.79 $\pm$ 0.17 \\
FreeDyG & 99.26 $\pm$ 0.01 & \textbf{99.48 $\pm$ 0.01} & 89.61 $\pm$ 0.19 & 92.15 $\pm$ 0.16 & 92.51 $\pm$ 0.05 & \textbf{94.91 $\pm$ 0.01} & 96.28 $\pm$ 0.11 \\
DyGMamba & 99.15 $\pm$ 0.02 & 99.32 $\pm$ 0.01 & 89.21 $\pm$ 0.08 & 93.35 $\pm$ 0.20 & 92.65 $\pm$ 0.12 & \underline{94.77 $\pm$ 0.01} & 95.91 $\pm$ 0.15 \\
RepeatMixer & 99.16 $\pm$ 0.02 & 99.22 $\pm$ 0.01 & 92.76 $\pm$ 0.10 & 94.14 $\pm$ 0.06 & 92.66 $\pm$ 0.07 & --- & 96.74 $\pm$ 0.08 \\
TIDFormer & \underline{99.30 $\pm$ 0.03} & \underline{99.38 $\pm$ 0.03} & 93.42 $\pm$ 0.15 & \textbf{94.60 $\pm$ 0.17} & \underline{93.17 $\pm$ 0.06} & 94.76 $\pm$ 0.06 & \textbf{97.44 $\pm$ 0.10} \\
TAMI-GraphMixer & 98.89 $\pm$ 0.00 & 98.84 $\pm$ 0.04 & 83.53 $\pm$ 0.04 & 88.13 $\pm$ 0.41 & 90.97 $\pm$ 0.06 & 93.41 $\pm$ 0.11 & 96.20 $\pm$ 0.21 \\
TAMI-DyGFormer & 99.25 $\pm$ 0.01 & 99.29 $\pm$ 0.00 & 88.49 $\pm$ 0.14 & 94.03 $\pm$ 0.08 & 92.66 $\pm$ 0.21 & 94.74 $\pm$ 0.03 & 96.72 $\pm$ 0.21 \\
EdgeBank & 90.37 $\pm$ 0.00 & 94.86 $\pm$ 0.00 & 57.97 $\pm$ 0.00 & 79.29 $\pm$ 0.00 & 83.53 $\pm$ 0.00 & 74.95 $\pm$ 0.00 & 76.20 $\pm$ 0.00 \\
\midrule
\method{} & 98.63 $\pm$ 0.02 & 99.16 $\pm$ 0.01 & 83.68 $\pm$ 0.07 & 92.37 $\pm$ 0.02 & \textbf{93.83 $\pm$ 0.02} & 94.11 $\pm$ 0.01 & 91.44 $\pm$ 0.22 \\
\bottomrule
\end{tabular}
\par\medskip
\begin{tabular}{@{}lcccccc@{}}
\toprule
Method & Flights & Can. Parl. & US Legis. & UN Trade & UN Vote & Contact \\
\midrule
JODIE & 95.60 $\pm$ 1.73 & 69.26 $\pm$ 0.31 & 75.05 $\pm$ 1.52 & 64.94 $\pm$ 0.31 & 63.91 $\pm$ 0.81 & 95.31 $\pm$ 1.33 \\
DyRep & 95.29 $\pm$ 0.72 & 66.54 $\pm$ 2.76 & 75.34 $\pm$ 0.39 & 63.21 $\pm$ 0.93 & 62.81 $\pm$ 0.80 & 95.98 $\pm$ 0.15 \\
TGN & 97.95 $\pm$ 0.14 & 70.88 $\pm$ 2.34 & 75.99 $\pm$ 0.58 & 65.03 $\pm$ 1.37 & 65.72 $\pm$ 2.17 & 96.89 $\pm$ 0.56 \\
CTAN & 93.64 $\pm$ 0.54 & 87.01 $\pm$ 1.10 & \underline{79.44 $\pm$ 2.28} & 69.01 $\pm$ 0.38 & 70.19 $\pm$ 1.26 & 98.53 $\pm$ 0.09 \\
CAWN & 98.51 $\pm$ 0.01 & 69.82 $\pm$ 2.34 & 70.58 $\pm$ 0.48 & 65.39 $\pm$ 0.12 & 52.84 $\pm$ 0.10 & 90.26 $\pm$ 0.28 \\
NAT & 98.57 $\pm$ 0.12 & 79.72 $\pm$ 1.76 & 78.71 $\pm$ 0.87 & 73.95 $\pm$ 1.16 & 70.45 $\pm$ 0.68 & 97.39 $\pm$ 0.22 \\
PINT & 98.80 $\pm$ 0.02 & 68.36 $\pm$ 1.43 & 74.85 $\pm$ 0.97 & 70.20 $\pm$ 0.58 & 66.25 $\pm$ 0.78 & 98.64 $\pm$ 0.02 \\
TPNet & 98.93 $\pm$ 0.02 & 90.28 $\pm$ 0.37 & \textbf{80.58 $\pm$ 0.23} & \textbf{87.24 $\pm$ 0.65} & \underline{75.12 $\pm$ 0.29} & 98.66 $\pm$ 0.01 \\
CNE-N & 98.92 $\pm$ 0.01 & 86.09 $\pm$ 0.14 & 78.69 $\pm$ 0.12 & 76.92 $\pm$ 0.06 & 66.40 $\pm$ 0.12 & 98.40 $\pm$ 0.02 \\
TAWRMAC & 98.84 $\pm$ 0.93 & 76.42 $\pm$ 1.47 & 73.89 $\pm$ 0.52 & 68.11 $\pm$ 0.90 & 66.50 $\pm$ 1.17 & \textbf{98.80 $\pm$ 0.03} \\
TGAT & 94.03 $\pm$ 0.18 & 70.73 $\pm$ 0.72 & 68.52 $\pm$ 3.16 & 61.47 $\pm$ 0.18 & 52.21 $\pm$ 0.98 & 96.28 $\pm$ 0.09 \\
TCL & 91.23 $\pm$ 0.02 & 68.67 $\pm$ 2.67 & 69.59 $\pm$ 0.48 & 62.21 $\pm$ 0.03 & 51.90 $\pm$ 0.30 & 92.44 $\pm$ 0.12 \\
GraphMixer & 90.99 $\pm$ 0.05 & 77.04 $\pm$ 0.46 & 70.74 $\pm$ 1.02 & 62.61 $\pm$ 0.27 & 52.11 $\pm$ 0.16 & 91.92 $\pm$ 0.03 \\
DyGFormer & 98.91 $\pm$ 0.01 & 97.36 $\pm$ 0.45 & 71.11 $\pm$ 0.59 & 66.46 $\pm$ 1.29 & 55.55 $\pm$ 0.42 & 98.29 $\pm$ 0.01 \\
FreeDyG & --- & --- & --- & --- & --- & --- \\
DyGMamba & \textbf{98.95 $\pm$ 0.05} & \textbf{99.57 $\pm$ 0.08} & 71.75 $\pm$ 0.26 & 67.50 $\pm$ 0.24 & 56.39 $\pm$ 0.18 & 98.43 $\pm$ 0.12 \\
RepeatMixer & --- & --- & --- & --- & --- & --- \\
TIDFormer & --- & --- & --- & --- & --- & --- \\
TAMI-GraphMixer & 96.75 $\pm$ 0.01 & 78.38 $\pm$ 0.07 & 71.57 $\pm$ 0.36 & 62.98 $\pm$ 0.49 & 57.74 $\pm$ 0.08 & 95.26 $\pm$ 0.06 \\
TAMI-DyGFormer & \underline{98.94 $\pm$ 0.00} & \underline{98.67 $\pm$ 0.12} & 71.40 $\pm$ 0.28 & 66.39 $\pm$ 0.04 & 56.02 $\pm$ 0.03 & 98.70 $\pm$ 0.01 \\
EdgeBank & 89.35 $\pm$ 0.00 & 64.55 $\pm$ 0.00 & 58.39 $\pm$ 0.00 & 60.41 $\pm$ 0.00 & 58.49 $\pm$ 0.00 & 92.58 $\pm$ 0.00 \\
\midrule
\method{} & 98.55 $\pm$ 0.01 & 84.33 $\pm$ 0.19 & 78.94 $\pm$ 0.09 & \underline{85.96 $\pm$ 0.08} & \textbf{77.74 $\pm$ 0.05} & \underline{98.79 $\pm$ 0.01} \\
\bottomrule
\end{tabular}

%% file: tables/rq1_historical_auc_full.tex
\begin{tabular}{@{}lccccccc@{}}
\toprule
Method & Wikipedia & Reddit & MOOC & LastFM & Enron & Social Evo. & UCI \\
\midrule
JODIE & 80.77 $\pm$ 0.73 & 80.52 $\pm$ 0.32 & 82.75 $\pm$ 0.83 & 75.22 $\pm$ 2.36 & 75.39 $\pm$ 2.37 & 90.06 $\pm$ 3.15 & 78.64 $\pm$ 3.50 \\
DyRep & 77.74 $\pm$ 0.33 & 80.15 $\pm$ 0.18 & 81.06 $\pm$ 0.94 & 74.65 $\pm$ 1.98 & 74.69 $\pm$ 3.55 & 93.12 $\pm$ 0.34 & 57.91 $\pm$ 3.12 \\
TGN & 82.74 $\pm$ 0.32 & 81.11 $\pm$ 0.19 & 88.00 $\pm$ 1.80 & 77.97 $\pm$ 3.04 & 77.09 $\pm$ 2.22 & 94.71 $\pm$ 0.53 & 77.25 $\pm$ 2.68 \\
CTAN & \underline{95.43 $\pm$ 0.07} & \underline{90.63 $\pm$ 2.28} & 73.87 $\pm$ 2.77 & 79.50 $\pm$ 0.82 & 81.95 $\pm$ 1.64 & --- & 75.05 $\pm$ 0.13 \\
CAWN & 67.84 $\pm$ 0.64 & 80.27 $\pm$ 0.30 & 71.57 $\pm$ 1.07 & 67.88 $\pm$ 0.24 & 65.10 $\pm$ 0.34 & 87.43 $\pm$ 0.15 & 57.86 $\pm$ 0.15 \\
NAT & 69.32 $\pm$ 2.78 & 79.36 $\pm$ 0.31 & 84.93 $\pm$ 2.89 & 75.89 $\pm$ 2.21 & 73.22 $\pm$ 2.18 & 91.87 $\pm$ 4.52 & 71.52 $\pm$ 1.61 \\
PINT & 89.25 $\pm$ 0.49 & 82.98 $\pm$ 0.63 & 87.44 $\pm$ 1.74 & 81.89 $\pm$ 0.96 & 83.80 $\pm$ 0.68 & 96.82 $\pm$ 0.24 & \underline{92.05 $\pm$ 0.36} \\
TPNet & 79.89 $\pm$ 2.47 & 81.87 $\pm$ 0.49 & \underline{93.45 $\pm$ 0.67} & \underline{84.64 $\pm$ 0.45} & 81.16 $\pm$ 1.28 & 97.22 $\pm$ 0.30 & 80.42 $\pm$ 0.64 \\
CNE-N & --- & --- & --- & --- & --- & --- & --- \\
TAWRMAC & 85.28 $\pm$ 0.40 & 82.31 $\pm$ 0.33 & 88.22 $\pm$ 4.37 & 78.99 $\pm$ 0.61 & 79.73 $\pm$ 0.48 & --- & 82.76 $\pm$ 0.31 \\
TGAT & 82.87 $\pm$ 0.22 & 79.33 $\pm$ 0.16 & 80.81 $\pm$ 0.67 & 64.27 $\pm$ 0.26 & 61.85 $\pm$ 1.43 & 93.08 $\pm$ 0.59 & 58.89 $\pm$ 1.57 \\
TCL & 85.76 $\pm$ 0.46 & 76.49 $\pm$ 0.16 & 72.09 $\pm$ 0.56 & 47.24 $\pm$ 3.13 & 67.95 $\pm$ 0.88 & 93.44 $\pm$ 0.68 & 72.25 $\pm$ 3.46 \\
GraphMixer & 87.68 $\pm$ 0.17 & 77.80 $\pm$ 0.12 & 76.68 $\pm$ 1.40 & 64.21 $\pm$ 0.73 & 75.27 $\pm$ 1.14 & 94.39 $\pm$ 0.31 & 77.54 $\pm$ 2.02 \\
DyGFormer & 78.80 $\pm$ 1.95 & 80.54 $\pm$ 0.29 & 87.04 $\pm$ 0.35 & 78.78 $\pm$ 0.35 & 76.55 $\pm$ 0.52 & 97.28 $\pm$ 0.07 & 76.97 $\pm$ 0.24 \\
FreeDyG & 82.78 $\pm$ 0.30 & 85.92 $\pm$ 0.10 & 88.32 $\pm$ 0.99 & 73.53 $\pm$ 0.12 & 75.74 $\pm$ 0.72 & \underline{97.42 $\pm$ 0.02} & 80.38 $\pm$ 0.26 \\
DyGMamba & 78.99 $\pm$ 1.24 & 81.71 $\pm$ 0.49 & 87.91 $\pm$ 0.93 & 79.82 $\pm$ 0.27 & 77.73 $\pm$ 0.61 & 97.27 $\pm$ 0.30 & 75.43 $\pm$ 1.99 \\
RepeatMixer & 86.23 $\pm$ 0.60 & 83.41 $\pm$ 0.21 & 92.68 $\pm$ 0.33 & 84.38 $\pm$ 0.21 & \underline{84.78 $\pm$ 0.29} & --- & 78.85 $\pm$ 0.43 \\
TIDFormer & --- & --- & --- & --- & --- & --- & --- \\
TAMI-GraphMixer & --- & --- & --- & --- & --- & --- & --- \\
TAMI-DyGFormer & --- & --- & --- & --- & --- & --- & --- \\
EdgeBank & 77.27 $\pm$ 0.00 & 78.58 $\pm$ 0.00 & 61.90 $\pm$ 0.00 & 78.09 $\pm$ 0.00 & 79.59 $\pm$ 0.00 & 85.81 $\pm$ 0.00 & 69.56 $\pm$ 0.00 \\
\midrule
\method{} & \textbf{96.35 $\pm$ 0.02} & \textbf{93.71 $\pm$ 0.01} & \textbf{98.15 $\pm$ 0.01} & \textbf{94.26 $\pm$ 0.01} & \textbf{94.96 $\pm$ 0.06} & \textbf{99.46 $\pm$ 0.00} & \textbf{98.11 $\pm$ 0.01} \\
\bottomrule
\end{tabular}
\par\medskip
\begin{tabular}{@{}lcccccc@{}}
\toprule
Method & Flights & Can. Parl. & US Legis. & UN Trade & UN Vote & Contact \\
\midrule
JODIE & 68.97 $\pm$ 1.87 & 62.44 $\pm$ 1.11 & 67.47 $\pm$ 6.40 & 68.92 $\pm$ 1.40 & 76.84 $\pm$ 1.01 & 96.35 $\pm$ 0.92 \\
DyRep & 69.43 $\pm$ 0.90 & 70.16 $\pm$ 1.70 & 91.44 $\pm$ 1.18 & 64.36 $\pm$ 1.40 & 74.72 $\pm$ 1.43 & 96.00 $\pm$ 0.23 \\
TGN & 68.39 $\pm$ 0.95 & 73.23 $\pm$ 3.08 & 83.53 $\pm$ 4.53 & 63.93 $\pm$ 5.41 & 73.40 $\pm$ 5.20 & 93.76 $\pm$ 1.29 \\
CTAN & \underline{78.87 $\pm$ 7.94} & 89.06 $\pm$ 1.14 & 89.88 $\pm$ 0.54 & 69.75 $\pm$ 0.90 & 78.62 $\pm$ 1.36 & \underline{98.37 $\pm$ 0.10} \\
CAWN & 66.11 $\pm$ 0.71 & 72.06 $\pm$ 3.94 & 78.62 $\pm$ 7.46 & 63.09 $\pm$ 0.74 & 51.27 $\pm$ 0.33 & 83.06 $\pm$ 0.32 \\
NAT & 67.11 $\pm$ 2.16 & 81.47 $\pm$ 2.58 & 94.63 $\pm$ 1.16 & 80.68 $\pm$ 0.98 & 76.47 $\pm$ 2.15 & 96.79 $\pm$ 0.50 \\
PINT & 69.52 $\pm$ 0.90 & 73.89 $\pm$ 4.50 & 83.45 $\pm$ 2.85 & 76.95 $\pm$ 1.92 & 76.61 $\pm$ 3.05 & 97.34 $\pm$ 0.16 \\
TPNet & 71.82 $\pm$ 0.82 & 86.39 $\pm$ 3.73 & \textbf{96.28 $\pm$ 0.44} & \underline{88.90 $\pm$ 1.00} & 78.43 $\pm$ 1.09 & 97.73 $\pm$ 0.11 \\
CNE-N & --- & --- & --- & --- & --- & --- \\
TAWRMAC & 69.27 $\pm$ 0.83 & 79.99 $\pm$ 1.66 & 92.95 $\pm$ 2.51 & 74.44 $\pm$ 0.83 & 79.15 $\pm$ 2.51 & 97.35 $\pm$ 0.10 \\
TGAT & 72.20 $\pm$ 0.16 & 70.86 $\pm$ 0.94 & 73.47 $\pm$ 5.25 & 60.37 $\pm$ 0.68 & 53.95 $\pm$ 3.15 & 95.39 $\pm$ 0.43 \\
TCL & 70.57 $\pm$ 0.18 & 69.95 $\pm$ 3.70 & 83.97 $\pm$ 3.71 & 61.43 $\pm$ 1.04 & 52.29 $\pm$ 2.39 & 93.34 $\pm$ 0.19 \\
GraphMixer & 70.37 $\pm$ 0.23 & 79.03 $\pm$ 1.01 & 85.17 $\pm$ 0.70 & 63.20 $\pm$ 1.54 & 52.61 $\pm$ 1.44 & 93.14 $\pm$ 0.34 \\
DyGFormer & 68.09 $\pm$ 0.43 & 97.61 $\pm$ 0.40 & 90.77 $\pm$ 1.96 & 73.86 $\pm$ 1.13 & 64.27 $\pm$ 1.78 & 97.17 $\pm$ 0.05 \\
FreeDyG & --- & --- & --- & --- & --- & --- \\
DyGMamba & 68.98 $\pm$ 0.81 & \textbf{99.82 $\pm$ 0.10} & 88.36 $\pm$ 1.78 & 71.41 $\pm$ 0.21 & 65.17 $\pm$ 1.24 & 97.27 $\pm$ 0.06 \\
RepeatMixer & --- & --- & --- & --- & --- & --- \\
TIDFormer & --- & --- & --- & --- & --- & --- \\
TAMI-GraphMixer & --- & --- & --- & --- & --- & --- \\
TAMI-DyGFormer & --- & --- & --- & --- & --- & --- \\
EdgeBank & 74.64 $\pm$ 0.00 & 63.04 $\pm$ 0.00 & 67.41 $\pm$ 0.00 & 86.61 $\pm$ 0.00 & \underline{89.62 $\pm$ 0.00} & 92.17 $\pm$ 0.00 \\
\midrule
\method{} & \textbf{94.85 $\pm$ 0.23} & \underline{98.18 $\pm$ 0.05} & \underline{95.88 $\pm$ 0.24} & \textbf{91.74 $\pm$ 0.18} & \textbf{91.09 $\pm$ 0.26} & \textbf{99.41 $\pm$ 0.00} \\
\bottomrule
\end{tabular}

%% file: tables/rq1_inductive_auc_full.tex
\begin{tabular}{@{}lccccccc@{}}
\toprule
Method & Wikipedia & Reddit & MOOC & LastFM & Enron & Social Evo. & UCI \\
\midrule
JODIE & 70.96 $\pm$ 0.78 & 83.51 $\pm$ 0.15 & 66.63 $\pm$ 2.30 & 61.32 $\pm$ 3.49 & 70.92 $\pm$ 1.05 & 90.01 $\pm$ 3.19 & 64.14 $\pm$ 1.26 \\
DyRep & 67.36 $\pm$ 0.96 & 82.90 $\pm$ 0.31 & 63.26 $\pm$ 1.01 & 62.15 $\pm$ 2.12 & 68.73 $\pm$ 1.34 & 93.07 $\pm$ 0.38 & 54.25 $\pm$ 2.01 \\
TGN & 80.97 $\pm$ 0.31 & 84.56 $\pm$ 0.24 & 77.44 $\pm$ 2.86 & 65.46 $\pm$ 4.27 & 71.34 $\pm$ 2.46 & 95.24 $\pm$ 0.56 & 64.11 $\pm$ 1.04 \\
CTAN & \underline{93.67 $\pm$ 0.11} & \underline{91.42 $\pm$ 2.18} & 68.17 $\pm$ 3.73 & \underline{78.70 $\pm$ 0.87} & 75.40 $\pm$ 1.92 & --- & 66.51 $\pm$ 0.25 \\
CAWN & 70.95 $\pm$ 0.95 & 88.04 $\pm$ 0.29 & 70.32 $\pm$ 1.43 & 67.92 $\pm$ 0.44 & 75.17 $\pm$ 0.50 & 89.93 $\pm$ 0.15 & 58.06 $\pm$ 0.26 \\
NAT & 52.11 $\pm$ 5.83 & 76.01 $\pm$ 1.04 & 75.56 $\pm$ 2.39 & 58.54 $\pm$ 2.37 & 69.62 $\pm$ 1.12 & 92.53 $\pm$ 4.20 & 49.73 $\pm$ 1.20 \\
PINT & 77.03 $\pm$ 1.93 & 81.02 $\pm$ 0.68 & 77.00 $\pm$ 2.24 & 69.51 $\pm$ 0.61 & 78.91 $\pm$ 0.96 & 97.10 $\pm$ 0.22 & \underline{79.42 $\pm$ 0.31} \\
TPNet & 75.36 $\pm$ 3.41 & 81.64 $\pm$ 0.42 & \underline{89.07 $\pm$ 0.63} & 72.48 $\pm$ 0.90 & 75.44 $\pm$ 1.38 & 97.59 $\pm$ 0.25 & 70.85 $\pm$ 0.96 \\
CNE-N & --- & --- & --- & --- & --- & --- & --- \\
TAWRMAC & 85.36 $\pm$ 1.67 & 85.14 $\pm$ 0.35 & 82.56 $\pm$ 2.26 & 67.83 $\pm$ 1.17 & 77.33 $\pm$ 0.46 & --- & 75.13 $\pm$ 1.03 \\
TGAT & 81.93 $\pm$ 0.22 & 87.13 $\pm$ 0.20 & 73.18 $\pm$ 0.33 & 63.99 $\pm$ 0.21 & 60.45 $\pm$ 2.12 & 92.94 $\pm$ 0.61 & 60.80 $\pm$ 1.01 \\
TCL & 82.19 $\pm$ 0.48 & 84.67 $\pm$ 0.29 & 70.36 $\pm$ 0.37 & 46.93 $\pm$ 2.59 & 67.64 $\pm$ 0.86 & 93.44 $\pm$ 0.72 & 70.05 $\pm$ 1.86 \\
GraphMixer & 84.28 $\pm$ 0.30 & 82.21 $\pm$ 0.13 & 72.45 $\pm$ 0.72 & 60.22 $\pm$ 0.32 & 71.53 $\pm$ 0.85 & 94.22 $\pm$ 0.32 & 74.59 $\pm$ 0.74 \\
DyGFormer & 75.09 $\pm$ 3.70 & 86.23 $\pm$ 0.51 & 80.76 $\pm$ 0.76 & 69.25 $\pm$ 0.36 & 74.07 $\pm$ 0.64 & 97.51 $\pm$ 0.06 & 65.96 $\pm$ 1.18 \\
FreeDyG & 82.74 $\pm$ 0.32 & 84.38 $\pm$ 0.21 & 78.47 $\pm$ 0.94 & 72.30 $\pm$ 0.59 & 77.27 $\pm$ 0.61 & \textbf{98.47 $\pm$ 0.02} & 75.39 $\pm$ 0.57 \\
DyGMamba & 75.64 $\pm$ 2.42 & 86.35 $\pm$ 0.52 & 81.08 $\pm$ 0.82 & 68.74 $\pm$ 0.55 & 75.47 $\pm$ 1.41 & 97.37 $\pm$ 0.26 & 66.83 $\pm$ 2.83 \\
RepeatMixer & 83.67 $\pm$ 0.89 & 86.85 $\pm$ 0.84 & 81.94 $\pm$ 0.95 & 67.97 $\pm$ 0.40 & \underline{80.53 $\pm$ 0.15} & --- & 76.56 $\pm$ 0.30 \\
TIDFormer & --- & --- & --- & --- & --- & --- & --- \\
TAMI-GraphMixer & --- & --- & --- & --- & --- & --- & --- \\
TAMI-DyGFormer & --- & --- & --- & --- & --- & --- & --- \\
EdgeBank & 81.73 $\pm$ 0.00 & 85.93 $\pm$ 0.00 & 48.18 $\pm$ 0.00 & 77.37 $\pm$ 0.00 & 75.00 $\pm$ 0.00 & 87.88 $\pm$ 0.00 & 58.03 $\pm$ 0.00 \\
\midrule
\method{} & \textbf{96.73 $\pm$ 0.03} & \textbf{98.97 $\pm$ 0.04} & \textbf{97.75 $\pm$ 0.02} & \textbf{97.40 $\pm$ 0.01} & \textbf{88.18 $\pm$ 0.64} & \underline{98.19 $\pm$ 0.15} & \textbf{88.56 $\pm$ 0.10} \\
\bottomrule
\end{tabular}
\par\medskip
\begin{tabular}{@{}lcccccc@{}}
\toprule
Method & Flights & Can. Parl. & US Legis. & UN Trade & UN Vote & Contact \\
\midrule
JODIE & 69.99 $\pm$ 3.10 & 52.88 $\pm$ 0.80 & 59.05 $\pm$ 5.52 & 66.82 $\pm$ 1.27 & 73.73 $\pm$ 1.61 & 94.47 $\pm$ 1.08 \\
DyRep & 71.13 $\pm$ 1.55 & 63.53 $\pm$ 0.65 & 89.44 $\pm$ 0.71 & 65.60 $\pm$ 1.28 & 72.80 $\pm$ 2.16 & 94.23 $\pm$ 0.18 \\
TGN & 71.63 $\pm$ 1.72 & 69.57 $\pm$ 2.81 & 78.12 $\pm$ 4.46 & 66.37 $\pm$ 5.39 & 72.69 $\pm$ 3.72 & 91.64 $\pm$ 1.72 \\
CTAN & 79.82 $\pm$ 7.80 & 85.99 $\pm$ 0.95 & 86.09 $\pm$ 1.11 & 69.74 $\pm$ 1.54 & 76.64 $\pm$ 1.02 & \underline{97.93 $\pm$ 0.12} \\
CAWN & 69.70 $\pm$ 0.75 & 72.93 $\pm$ 1.78 & 76.45 $\pm$ 7.02 & 71.73 $\pm$ 0.74 & 52.75 $\pm$ 0.90 & 87.68 $\pm$ 0.24 \\
NAT & 59.73 $\pm$ 1.61 & 80.03 $\pm$ 1.75 & \underline{93.04 $\pm$ 1.35} & 80.77 $\pm$ 1.31 & 77.36 $\pm$ 1.52 & 92.51 $\pm$ 1.16 \\
PINT & 59.61 $\pm$ 1.65 & 65.30 $\pm$ 6.98 & 79.09 $\pm$ 3.45 & 75.76 $\pm$ 1.54 & 74.11 $\pm$ 2.51 & 94.71 $\pm$ 0.33 \\
TPNet & 64.21 $\pm$ 1.30 & 85.05 $\pm$ 2.71 & \textbf{94.48 $\pm$ 0.50} & \underline{89.56 $\pm$ 0.87} & \underline{79.13 $\pm$ 1.16} & 95.93 $\pm$ 0.23 \\
CNE-N & --- & --- & --- & --- & --- & --- \\
TAWRMAC & 63.82 $\pm$ 1.91 & 73.96 $\pm$ 1.77 & 89.81 $\pm$ 2.45 & 67.60 $\pm$ 2.45 & 75.31 $\pm$ 2.45 & 95.87 $\pm$ 0.11 \\
TGAT & 73.47 $\pm$ 0.18 & 72.47 $\pm$ 1.18 & 71.62 $\pm$ 5.42 & 66.13 $\pm$ 0.78 & 53.04 $\pm$ 2.58 & 94.10 $\pm$ 0.41 \\
TCL & 72.54 $\pm$ 0.19 & 69.47 $\pm$ 2.12 & 82.54 $\pm$ 3.91 & 67.80 $\pm$ 1.21 & 52.02 $\pm$ 1.64 & 91.23 $\pm$ 0.19 \\
GraphMixer & 72.21 $\pm$ 0.21 & 70.52 $\pm$ 0.94 & 84.22 $\pm$ 0.91 & 66.53 $\pm$ 1.22 & 51.89 $\pm$ 0.74 & 90.96 $\pm$ 0.27 \\
DyGFormer & 69.53 $\pm$ 1.17 & \underline{96.70 $\pm$ 0.59} & 87.96 $\pm$ 1.80 & 62.56 $\pm$ 1.51 & 53.37 $\pm$ 1.26 & 95.01 $\pm$ 0.15 \\
FreeDyG & --- & --- & --- & --- & --- & --- \\
DyGMamba & 71.16 $\pm$ 3.24 & \textbf{99.56 $\pm$ 0.21} & 86.06 $\pm$ 2.27 & 67.60 $\pm$ 0.64 & 54.09 $\pm$ 0.06 & 95.68 $\pm$ 0.20 \\
RepeatMixer & --- & --- & --- & --- & --- & --- \\
TIDFormer & --- & --- & --- & --- & --- & --- \\
TAMI-GraphMixer & --- & --- & --- & --- & --- & --- \\
TAMI-DyGFormer & --- & --- & --- & --- & --- & --- \\
EdgeBank & \underline{81.10 $\pm$ 0.00} & 61.41 $\pm$ 0.00 & 68.66 $\pm$ 0.00 & 74.20 $\pm$ 0.00 & 72.85 $\pm$ 0.00 & 85.87 $\pm$ 0.00 \\
\midrule
\method{} & \textbf{98.71 $\pm$ 0.21} & 83.06 $\pm$ 0.23 & 84.51 $\pm$ 0.22 & \textbf{96.46 $\pm$ 0.16} & \textbf{94.45 $\pm$ 0.02} & \textbf{99.74 $\pm$ 0.00} \\
\bottomrule
\end{tabular}

%% file: tables/rq1_random_auc_full.tex
\begin{tabular}{@{}lccccccc@{}}
\toprule
Method & Wikipedia & Reddit & MOOC & LastFM & Enron & Social Evo. & UCI \\
\midrule
JODIE & 96.33 $\pm$ 0.07 & 98.31 $\pm$ 0.05 & 83.81 $\pm$ 2.09 & 70.49 $\pm$ 1.66 & 87.96 $\pm$ 0.52 & 92.05 $\pm$ 0.46 & 90.44 $\pm$ 0.49 \\
DyRep & 94.37 $\pm$ 0.09 & 98.17 $\pm$ 0.05 & 85.03 $\pm$ 0.58 & 71.16 $\pm$ 1.89 & 84.89 $\pm$ 3.00 & 90.76 $\pm$ 0.21 & 68.77 $\pm$ 2.34 \\
TGN & 98.37 $\pm$ 0.07 & 98.60 $\pm$ 0.06 & 91.21 $\pm$ 1.15 & 78.47 $\pm$ 2.94 & 88.32 $\pm$ 0.99 & 95.39 $\pm$ 0.17 & 92.03 $\pm$ 1.13 \\
CTAN & 97.00 $\pm$ 0.21 & 97.24 $\pm$ 0.75 & 85.40 $\pm$ 2.67 & 85.12 $\pm$ 0.77 & 87.09 $\pm$ 1.51 & --- & 76.25 $\pm$ 2.83 \\
CAWN & 98.54 $\pm$ 0.04 & 99.01 $\pm$ 0.01 & 80.38 $\pm$ 0.26 & 85.92 $\pm$ 0.10 & 90.45 $\pm$ 0.14 & 87.34 $\pm$ 0.08 & 93.87 $\pm$ 0.08 \\
NAT & 97.75 $\pm$ 0.11 & 99.09 $\pm$ 0.10 & 87.42 $\pm$ 0.58 & 86.92 $\pm$ 2.72 & 91.68 $\pm$ 0.83 & 90.84 $\pm$ 3.72 & 92.31 $\pm$ 0.37 \\
PINT & 98.16 $\pm$ 0.06 & 99.09 $\pm$ 0.03 & 90.55 $\pm$ 0.43 & 89.28 $\pm$ 1.63 & 92.87 $\pm$ 0.34 & 96.16 $\pm$ 0.02 & 95.57 $\pm$ 0.16 \\
TPNet & \underline{99.30 $\pm$ 0.02} & 99.22 $\pm$ 0.00 & \textbf{97.17 $\pm$ 0.08} & \textbf{94.39 $\pm$ 0.04} & 93.98 $\pm$ 0.26 & \underline{96.43 $\pm$ 0.02} & \textbf{96.79 $\pm$ 0.05} \\
CNE-N & 99.01 $\pm$ 0.05 & 99.15 $\pm$ 0.01 & \underline{95.69 $\pm$ 0.07} & 93.13 $\pm$ 0.11 & 93.12 $\pm$ 0.09 & 96.24 $\pm$ 0.14 & \underline{96.03 $\pm$ 0.08} \\
TAWRMAC & \underline{99.30 $\pm$ 0.01} & \underline{99.27 $\pm$ 0.01} & 92.57 $\pm$ 0.60 & 90.17 $\pm$ 0.14 & 93.97 $\pm$ 0.48 & --- & 95.40 $\pm$ 0.22 \\
TGAT & 96.67 $\pm$ 0.07 & 98.47 $\pm$ 0.02 & 87.11 $\pm$ 0.19 & 71.59 $\pm$ 0.18 & 68.89 $\pm$ 1.10 & 94.76 $\pm$ 0.16 & 78.53 $\pm$ 0.74 \\
TCL & 95.84 $\pm$ 0.18 & 97.42 $\pm$ 0.02 & 83.12 $\pm$ 0.18 & 64.06 $\pm$ 1.16 & 75.74 $\pm$ 0.72 & 94.84 $\pm$ 0.17 & 87.82 $\pm$ 1.36 \\
GraphMixer & 96.92 $\pm$ 0.03 & 97.17 $\pm$ 0.02 & 84.01 $\pm$ 0.17 & 73.53 $\pm$ 0.12 & 84.38 $\pm$ 0.21 & 95.23 $\pm$ 0.07 & 91.81 $\pm$ 0.67 \\
DyGFormer & 98.91 $\pm$ 0.02 & 99.15 $\pm$ 0.01 & 87.91 $\pm$ 0.58 & 93.05 $\pm$ 0.10 & 93.33 $\pm$ 0.13 & 96.30 $\pm$ 0.01 & 94.49 $\pm$ 0.26 \\
FreeDyG & \textbf{99.41 $\pm$ 0.01} & \textbf{99.50 $\pm$ 0.01} & 89.93 $\pm$ 0.35 & 93.42 $\pm$ 0.15 & \underline{94.01 $\pm$ 0.11} & \textbf{96.59 $\pm$ 0.04} & 95.00 $\pm$ 0.21 \\
DyGMamba & 99.08 $\pm$ 0.02 & \underline{99.27 $\pm$ 0.01} & 89.58 $\pm$ 0.12 & 93.31 $\pm$ 0.18 & 93.34 $\pm$ 0.23 & 96.38 $\pm$ 0.02 & 94.77 $\pm$ 0.18 \\
RepeatMixer & 99.04 $\pm$ 0.01 & 99.15 $\pm$ 0.01 & 93.60 $\pm$ 0.39 & \underline{94.27 $\pm$ 0.02} & 93.47 $\pm$ 0.17 & --- & 95.36 $\pm$ 0.49 \\
TIDFormer & --- & --- & --- & --- & --- & --- & --- \\
TAMI-GraphMixer & --- & --- & --- & --- & --- & --- & --- \\
TAMI-DyGFormer & --- & --- & --- & --- & --- & --- & --- \\
EdgeBank & 90.78 $\pm$ 0.00 & 95.37 $\pm$ 0.00 & 60.86 $\pm$ 0.00 & 83.77 $\pm$ 0.00 & 87.05 $\pm$ 0.00 & 81.60 $\pm$ 0.00 & 77.30 $\pm$ 0.00 \\
\midrule
\method{} & 98.50 $\pm$ 0.02 & 99.06 $\pm$ 0.00 & 84.83 $\pm$ 0.07 & 92.58 $\pm$ 0.01 & \textbf{94.85 $\pm$ 0.05} & 95.85 $\pm$ 0.01 & 92.24 $\pm$ 0.07 \\
\bottomrule
\end{tabular}
\par\medskip
\begin{tabular}{@{}lcccccc@{}}
\toprule
Method & Flights & Can. Parl. & US Legis. & UN Trade & UN Vote & Contact \\
\midrule
JODIE & 96.21 $\pm$ 1.42 & 78.21 $\pm$ 0.23 & 82.85 $\pm$ 1.07 & 69.62 $\pm$ 0.44 & 68.53 $\pm$ 0.95 & 96.66 $\pm$ 0.89 \\
DyRep & 95.95 $\pm$ 0.62 & 73.35 $\pm$ 3.67 & 82.28 $\pm$ 0.32 & 67.44 $\pm$ 0.83 & 67.18 $\pm$ 1.04 & 96.48 $\pm$ 0.14 \\
TGN & 98.22 $\pm$ 0.13 & 76.99 $\pm$ 1.80 & 83.34 $\pm$ 0.43 & 69.10 $\pm$ 1.67 & 69.71 $\pm$ 2.65 & 97.54 $\pm$ 0.35 \\
CTAN & 94.57 $\pm$ 0.57 & 86.77 $\pm$ 1.17 & \underline{86.01 $\pm$ 1.68} & 72.00 $\pm$ 0.24 & 73.91 $\pm$ 1.08 & 98.88 $\pm$ 0.06 \\
CAWN & 98.45 $\pm$ 0.01 & 75.70 $\pm$ 3.27 & 77.16 $\pm$ 0.39 & 68.54 $\pm$ 0.18 & 53.09 $\pm$ 0.22 & 89.99 $\pm$ 0.34 \\
NAT & 98.69 $\pm$ 0.10 & 84.04 $\pm$ 1.13 & 85.36 $\pm$ 0.52 & 77.61 $\pm$ 1.36 & 75.32 $\pm$ 0.63 & 97.79 $\pm$ 0.16 \\
PINT & 98.89 $\pm$ 0.02 & 77.96 $\pm$ 1.46 & 82.10 $\pm$ 0.85 & 74.87 $\pm$ 0.53 & 70.69 $\pm$ 1.02 & 98.90 $\pm$ 0.02 \\
TPNet & \textbf{99.00 $\pm$ 0.02} & 92.05 $\pm$ 0.34 & \textbf{86.49 $\pm$ 0.18} & \textbf{89.17 $\pm$ 0.46} & \underline{79.88 $\pm$ 0.30} & 98.91 $\pm$ 0.01 \\
CNE-N & 98.96 $\pm$ 0.02 & 89.71 $\pm$ 0.07 & 84.14 $\pm$ 0.63 & 78.57 $\pm$ 0.07 & 69.52 $\pm$ 0.34 & 98.77 $\pm$ 0.01 \\
TAWRMAC & 98.93 $\pm$ 0.82 & 82.39 $\pm$ 1.40 & 80.23 $\pm$ 1.11 & 72.28 $\pm$ 1.40 & 70.70 $\pm$ 1.35 & \textbf{99.06 $\pm$ 0.02} \\
TGAT & 94.13 $\pm$ 0.17 & 75.69 $\pm$ 0.78 & 75.84 $\pm$ 1.99 & 64.01 $\pm$ 0.12 & 52.83 $\pm$ 1.12 & 96.95 $\pm$ 0.08 \\
TCL & 91.21 $\pm$ 0.02 & 72.46 $\pm$ 3.23 & 76.27 $\pm$ 0.63 & 64.72 $\pm$ 0.05 & 51.88 $\pm$ 0.36 & 94.15 $\pm$ 0.09 \\
GraphMixer & 91.13 $\pm$ 0.01 & 83.17 $\pm$ 0.53 & 76.96 $\pm$ 0.79 & 65.52 $\pm$ 0.51 & 52.46 $\pm$ 0.27 & 93.94 $\pm$ 0.02 \\
DyGFormer & 98.93 $\pm$ 0.01 & \underline{97.76 $\pm$ 0.41} & 77.90 $\pm$ 0.58 & 70.20 $\pm$ 1.44 & 57.12 $\pm$ 0.62 & 98.53 $\pm$ 0.01 \\
FreeDyG & --- & --- & --- & --- & --- & --- \\
DyGMamba & \underline{98.98 $\pm$ 0.05} & \textbf{99.69 $\pm$ 0.06} & 79.03 $\pm$ 0.26 & 71.41 $\pm$ 0.21 & 58.48 $\pm$ 0.12 & 98.68 $\pm$ 0.02 \\
RepeatMixer & --- & --- & --- & --- & --- & --- \\
TIDFormer & --- & --- & --- & --- & --- & --- \\
TAMI-GraphMixer & --- & --- & --- & --- & --- & --- \\
TAMI-DyGFormer & --- & --- & --- & --- & --- & --- \\
EdgeBank & 90.23 $\pm$ 0.00 & 64.14 $\pm$ 0.00 & 62.57 $\pm$ 0.00 & 66.75 $\pm$ 0.00 & 62.97 $\pm$ 0.00 & 94.34 $\pm$ 0.00 \\
\midrule
\method{} & 98.54 $\pm$ 0.01 & 89.50 $\pm$ 0.22 & 85.71 $\pm$ 0.07 & \underline{88.32 $\pm$ 0.03} & \textbf{82.51 $\pm$ 0.03} & \underline{99.05 $\pm$ 0.00} \\
\bottomrule
\end{tabular}

%% file: tables/rq3_all.tex
\begin{tabular}{@{}lccccccccccccc@{}}
\toprule
Random & Wiki. & Reddit & MOOC & LastFM & Enron & S.Evo & UCI & Flights & CanP. & USLeg. & UNTr. & UNVo. & Cont. \\
\midrule
LightGBM & 98.12 & 98.95 & 82.44 & 91.75 & 93.19 & 93.31 & 82.49 & 96.63 & \textbf{85.56} & 78.12 & 85.60 & 76.93 & 98.58 \\
HGB & 97.91 & 98.95 & 81.75 & 91.61 & 92.84 & --- & 83.81 & --- & --- & --- & --- & --- & --- \\
TabICL & 98.62 & 99.15 & 82.43 & 92.36 & 93.51 & --- & 83.49 & --- & --- & --- & --- & --- & --- \\
TabPFN & \textbf{98.63} & \textbf{99.16} & \textbf{83.68} & \textbf{92.37} & \textbf{93.83} & \textbf{94.11} & \textbf{91.44} & \textbf{98.55} & 84.33 & \textbf{78.94} & \textbf{85.96} & \textbf{77.74} & \textbf{98.79} \\
\bottomrule
\end{tabular}
\par\smallskip
\begin{tabular}{@{}lccccccccccccc@{}}
\toprule
Historical & Wiki. & Reddit & MOOC & LastFM & Enron & S.Evo & UCI & Flights & CanP. & USLeg. & UNTr. & UNVo. & Cont. \\
\midrule
LightGBM & 96.20 & 93.46 & 98.40 & 94.17 & 94.27 & 99.48 & 98.13 & 94.12 & \textbf{98.32} & \textbf{96.25} & 90.35 & \textbf{94.35} & 99.41 \\
HGB & 96.09 & 93.31 & 98.37 & 94.05 & 94.17 & --- & 98.08 & --- & --- & --- & --- & --- & --- \\
TabICL & 96.37 & \textbf{93.98} & 98.46 & 94.21 & 93.62 & --- & \textbf{98.36} & --- & --- & --- & --- & --- & --- \\
TabPFN & \textbf{96.65} & 93.78 & \textbf{98.48} & \textbf{94.48} & \textbf{94.40} & \textbf{99.50} & 98.34 & \textbf{94.96} & 98.02 & 95.26 & \textbf{91.34} & 90.46 & \textbf{99.46} \\
\bottomrule
\end{tabular}
\par\smallskip
\begin{tabular}{@{}lccccccccccccc@{}}
\toprule
Inductive & Wiki. & Reddit & MOOC & LastFM & Enron & S.Evo & UCI & Flights & CanP. & USLeg. & UNTr. & UNVo. & Cont. \\
\midrule
LightGBM & 95.90 & 98.60 & 98.05 & 97.60 & 82.25 & 95.19 & 90.19 & 96.58 & 74.60 & 74.14 & 84.85 & 85.82 & 99.49 \\
HGB & 95.76 & 98.49 & 97.97 & 97.55 & 82.15 & --- & 89.77 & --- & --- & --- & --- & --- & --- \\
TabICL & 95.90 & 98.76 & \textbf{98.23} & 97.86 & 85.48 & --- & \textbf{90.78} & --- & --- & --- & --- & --- & --- \\
TabPFN & \textbf{97.39} & \textbf{99.13} & 98.21 & \textbf{98.00} & \textbf{89.49} & \textbf{97.81} & 90.73 & \textbf{98.87} & \textbf{83.16} & \textbf{83.09} & \textbf{95.33} & \textbf{92.38} & \textbf{99.79} \\
\bottomrule
\end{tabular}